\documentclass[10pt,twocolumn,letterpaper]{article}

\usepackage{iccv}
\usepackage{times}
\usepackage{epsfig}
\usepackage{graphicx}
\usepackage{amsmath}
\usepackage{amssymb}
\usepackage{caption}

\usepackage{subfigure}
\usepackage{multirow}
\usepackage{wrapfig}

\usepackage{mathtools, cuted}
\usepackage{lipsum}

\usepackage[pagebackref=true,breaklinks=true,letterpaper=true,colorlinks,bookmarks=false]{hyperref}

\iccvfinalcopy 

\ificcvfinal\fi

\begin{document}

\title{MEDOE: A Multi-Expert Decoder and Output \\ Ensemble Framework for Long-tailed Semantic Segmentation}

\author{Junao Shen\textsuperscript{\rm 1}, Long Cheng\textsuperscript{\rm 2}, Kun Kuang\textsuperscript{\rm 1}, Fei Wu\textsuperscript{\rm 1}, Tian Feng\textsuperscript{\rm 1}\thanks{Corresponding authors}, Wei Zhang\textsuperscript{\rm 1} \\
\textsuperscript{\rm 1}Zhejiang University, Hangzhou, China \\
\textsuperscript{\rm 2}The Hong Kong University of Science and Technology, HongKong, China\\
{\tt\small \{jashen, wufei, t.feng, cstzhangwei\}@zju.edu.cn, longchen@ust.hk, kkun2010@gmail.com}
}

\newcommand{\fix}{\marginpar{FIX}}
\newcommand{\new}{\marginpar{NEW}}

\maketitle
\ificcvfinal\thispagestyle{empty}\fi

\begin{abstract}
   
  Long-tailed distribution of semantic categories, which has been often ignored in conventional methods, causes unsatisfactory performance in semantic segmentation on tail categories. In this paper, we focus on the problem of long-tailed semantic segmentation. Although some long-tailed recognition methods (\eg, re-sampling/re-weighting) have been proposed in other problems, they can probably compromise crucial contextual information and are thus hardly adaptable to the problem of long-tailed semantic segmentation. To address this issue, we propose MEDOE, a novel framework for long-tailed semantic segmentation via contextual information ensemble-and-grouping. The proposed two-sage framework comprises a multi-expert decoder (MED) and a multi-expert output ensemble (MOE). Specifically, the MED includes several ``experts". Based on the pixel frequency distribution, each expert takes the dataset masked according to the specific categories as input and generates contextual information self-adaptively for classification; The MOE adopts learnable decision weights for the ensemble of the experts' outputs. As a model-agnostic framework, our MEDOE can be flexibly and efficiently coupled with various popular deep neural networks (\eg, DeepLabv3+, OCRNet, and PSPNet) to improve their performance in long-tailed semantic segmentation. Experimental results show that the proposed framework outperforms the current methods on both Cityscapes and ADE20K datasets by up to 1.78\% in mIoU and 5.89\% in mAcc.

\end{abstract}

\section{Introduction} \label{intro}


Semantic segmentation aims to predict the semantic category for each pixel in an image. As a fundamental task in computer vision, semantic segmentation is crucial to various real-world applications (\eg, clinical analysis and automatic driving). Conventional methods follow the \emph{feature extractor \& context module} architecture. Especially, the context module contributes significantly because it can enhance the extraction of surrounding pixels and global information through large-scale field convolution or attention mechanisms~\cite{r11-deeplab,r13,r14,r15-ocr,r16-pspnet}. Such modifications have enabled the recent state-of-the-art methods to improve the performance in semantic segmentation on various benchmark datasets ~\cite{r1,r2,r39-ade20k,r3}. 


\begin{figure}[t]
\label{improvements}
    \centering
    \subfigure[Performance and pixel frequency on Cityscapes]{
    \centering
    \includegraphics[height = 1.1in, width = 2.8in]{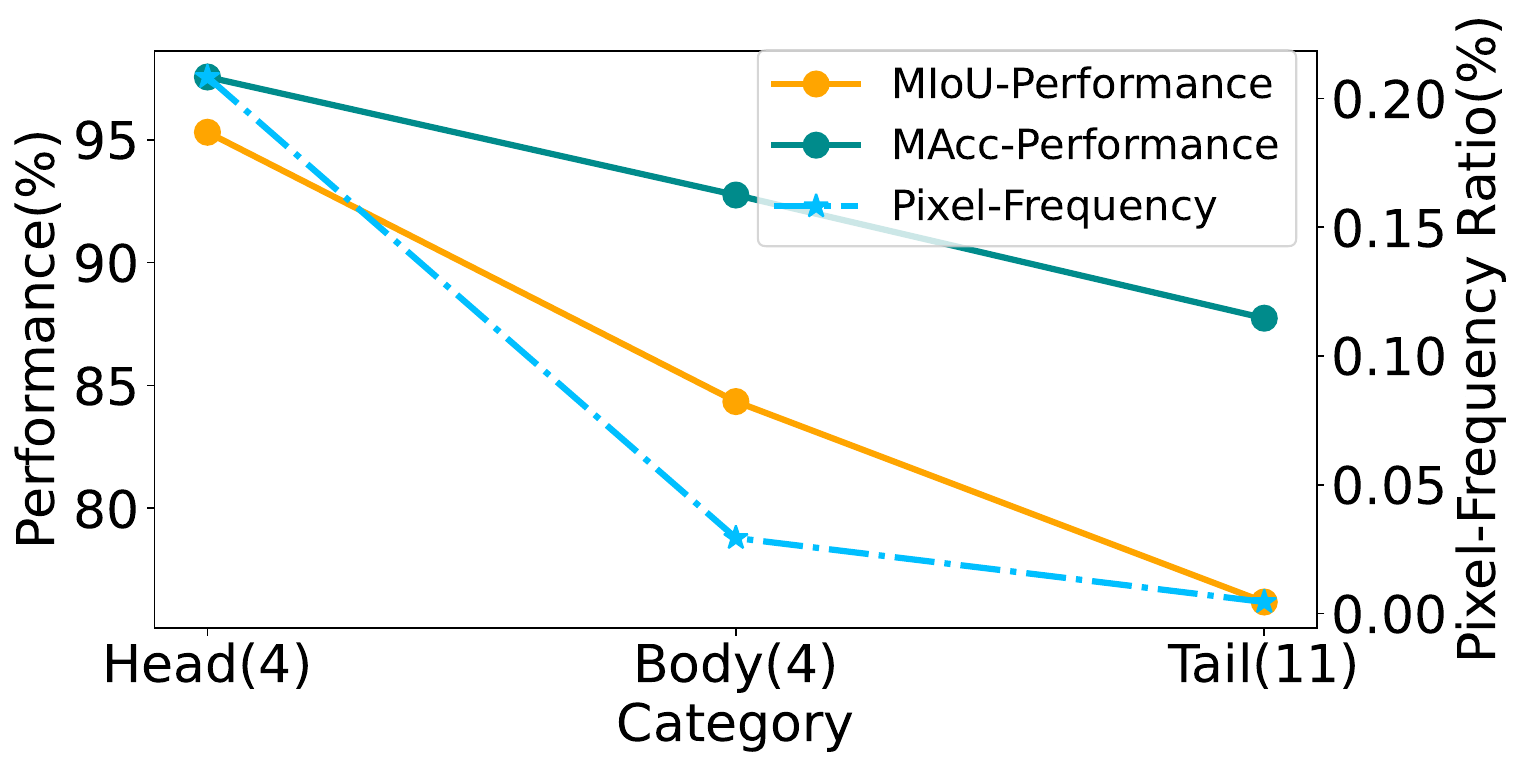}
    }
    \vspace{-1em}
    \subfigure[Performance and pixel frequency on ADE20K]{
    \centering
    \includegraphics[height = 1.1in, width = 2.8in]{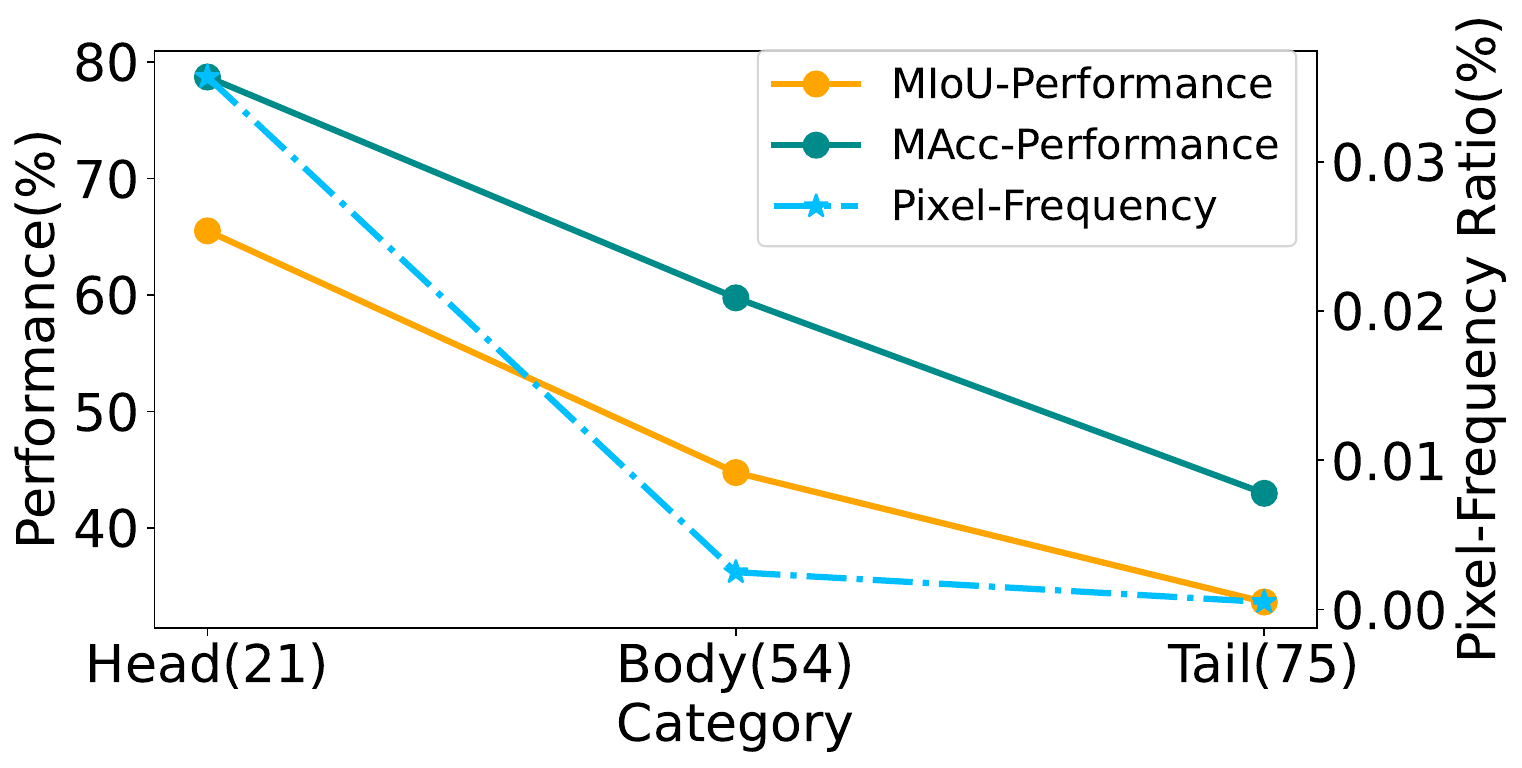}
    }
    \caption{Comparison of results in terms of mIoU~(\%) and mAcc~(\%) for head, body, and tail categories on Cityscapes~\cite{r38-2016cityscapes} and ADE20K ~\cite{r39-ade20k} for semantic segmantation using DeepLabv3+~\cite{r11-deeplab} (with ResNet-50c~\cite{he2016deep}). $(\cdot)$ denotes the categories' number in each term.}
    \label{fig_split}
    \vspace{-2em}
\end{figure}

\begin{figure}[t]
    \centering
    \subfigure[ Street Image]{
    \centering
    \includegraphics[height = 0.5in, width = 1in]{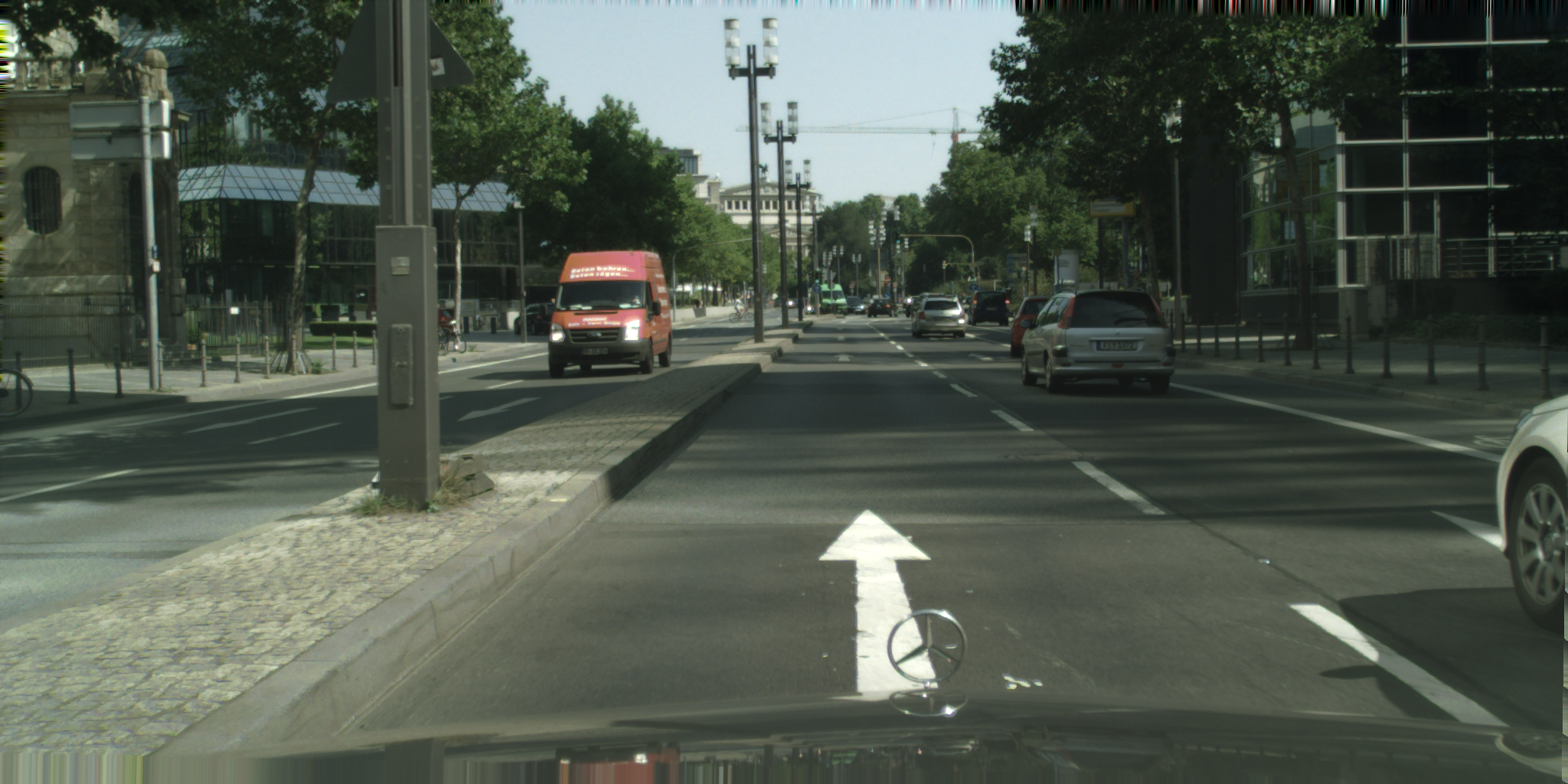}
    }
    \subfigure[Ground Truth]{
    \centering
    \includegraphics[height = 0.5in, width = 1in]{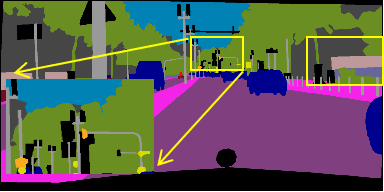}
    }
    \subfigure[Segmentation]{
    \centering
    \includegraphics[height = 0.5in, width = 1in]{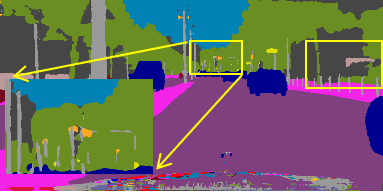}
    }
    \vspace{-1.25em}
    \caption{Segmentation example on Cityscapes, where tail categories (\eg, ``pole" and ``wall") are not well segmented. \label{fig_1}}
    \vspace{-1.5em}
\end{figure}

Despite the overall impressive performance, the above-mentioned methods still face challenges in semantic segmentation regarding data distribution. For example, Figure~\ref{fig_split} shows the segmentation performance for \emph{head}, \emph{body}, and \emph{tail} categories (\ie, categories ranked from top to bottom by pixel frequency) on two datasets, where positive correlations can be found between the data distribution and results. In particular, the performance descends for tail categories. This suggests the issue of \emph{long-tailed distribution} in semantic segmentation: a few \emph{head} categories relate to the majority of pixels, whereas many \emph{tail} categories correspond to significantly fewer pixels. Since a long-tailed distribution exists, processing all categories in the same pattern may lead to an excessive influence of head categories on the training and a negative impact on learning the contextual information about tail ones, which results in unsatisfactory per-pixel classification. Therefore, solving the problem of long-tailed semantic segmentation is critical to real-world applications. As shown in Figure~\ref{fig_1}, accurately distinguishing the ``poles" from street scene images is likely to prevent potential traffic accidents.


The most popular solution to the long-tailed recognition problem is to re-balance the contribution of each category in the training phase. Such methods~\cite{r17,r18,r29-lvis} can be further categorised into \emph{re-weighting}, \emph{re-sampling}.
1) Re-weighting methods increase the weights of tail categories, while decreasing those of head ones~\cite{r30,r31-oltr}, assuming that images are nearly independent and identically distributed (\ie, i.i.d.) to address the imbalance of the training set. This assumption enables the classification accuracy for each category to rely on the frequency of the corresponding images~\cite{r30,r32-RR}. However, re-weighting methods cannot serve per-pixel classification, for each pixel is usually highly correlated to the surrounding ones (\ie, not i.i.d.) given the contextual information in the image~\cite{r49-focal}. Re-weighting may also cause a heavy \emph{see-saw} phenomenon: the accuracy for head categories is compromised whereas tail categories are on purpose emphasized. 2) Re-sampling methods execute under-sampling for head categories and over-sampling (or even data augmentation) for tail categories~\cite{r20,r33,r34}. It is noteworthy that the random sampling strategy to ensure fairness leads to pixels independent from the image, which undermines the image's contextual information and is detrimental to semantic segmentation.

To avoid the disadvantages of re-weighting and re-sampling methods, recent studies have proposed a novel strategy called \emph{ensemble-and-grouping}. In particular, ensemble-and-grouping methods adopt a feature extractor trained on the entire imbalanced dataset for representation learning, and adjust the margins of classifiers using a multi-expert framework~\cite{r35-lfme,r36,r37-bbn} for re-balancing~\cite{r23-decoupling}. These methods, however, rely heavily on re-balancing, where the re-adjusting classifier is under-adapted to semantic segmentation due to ignoring the difference among head, body, and tail categories in contextual information.

Considering the aforementioned methods can hardly be directly extended to solve the long-tailed semantic segmentation problem, we propose MEDOE, a two-stage multi-expert decoder and output ensemble framework. Specifically, at the first stage, the feature map extracted by a backbone trained on the whole of an imbalanced dataset represents the elementary knowledge and is passed to a multi-expert decoder (MED). Each \emph{expert} (\ie, a pair of context module and classifier head) in the MED works on the specific dataset by \emph{expert-specific label-masking strategy}. In this strategy, the labels for certain categories (\ie, head or body categories) in each ground truth are masked. Together with the loss function, this strategy enables the experts to reduce the impact of head categories and irrelevant categories and acquire the contextual information of body or tail categories. At the second stage, a multi-expert output ensemble (MOE) is designed to group the outputs of all experts from the previous stage using decision weights. Instead of being user-specified, the weights are learned by a decision-maker to avoid the negative impact of over-confidence for tail categories. The proposed framework is model-agnostic and can be integrated with popular semantic segmentation methods, such as DeepLabv3+~\cite{r11-deeplab}, PSPNet~\cite{r16-pspnet}, OCRNet~\cite{r15-ocr}, and SegFormer~\cite{xie2021segformer}.

 To appropriately evaluate the performance of long-tailed semantic segmentation, we conduct extensive experiments with our MEDOE, especially for tail categories. Specifically, we extend the conventional setting in semantic segmentation by following long-tailed recognition setting~\cite{r30,r31-oltr} to test the robustness of the proposed framework with uniformly distributed data, taking into account the variation of real-world data distribution, (\ie, more person during rush hours than normal times). Experimental results suggest our MEDOE's effectiveness and robustness. Compared with the previous methods, the proposed framework is fairer to tail and body categories and thus can be applied to extensive data distributions. Besides, we employ a theoretical analysis to reveal that long-tailed recognition setting can reduce the effect of misclassifying the pixels of head categories as tail categories on the mIoU for tail categories, which enables a fair comparison of the performance for each category. 
 Although the information required by the expert-specific label-masking strategy is unavailable in the inference phase of semantic segmentation, we examine the ideal upper bounds of both MED and MOE to demonstrate the potential of the proposed framework.
The contributions of our work are summarized as,
\begin{itemize}
\vspace{-0.5em}
\itemsep-0.25em



\item an empirical study on the issue of long-tailed distribution in semantic segmentation, revealing its significance;

\item a model-agnostic framework based on the multi-expert decoder and output ensemble, outperforming several popular methods up to 1.78\% in mIoU and 5.89\% in mAcc on Cityscapes~\cite{r38-2016cityscapes} and ADE20K~\cite{r39-ade20k} datasets;

\item extending the task setting to evaluate the robustness long-tailed semantic segmentation performance for body and tail categories; and

\item demonstrating the ideal information referred to \textit{Oracle} improves substantially the results, with an average gain of 5\% in mIoU and 6\% in mAcc on Cityscapes and 12\% in mIoU and 19\% in mAcc on ADE20K datasets.

\end{itemize}


\begin{figure*}[t]
    \centering
\includegraphics[width=0.8\linewidth]{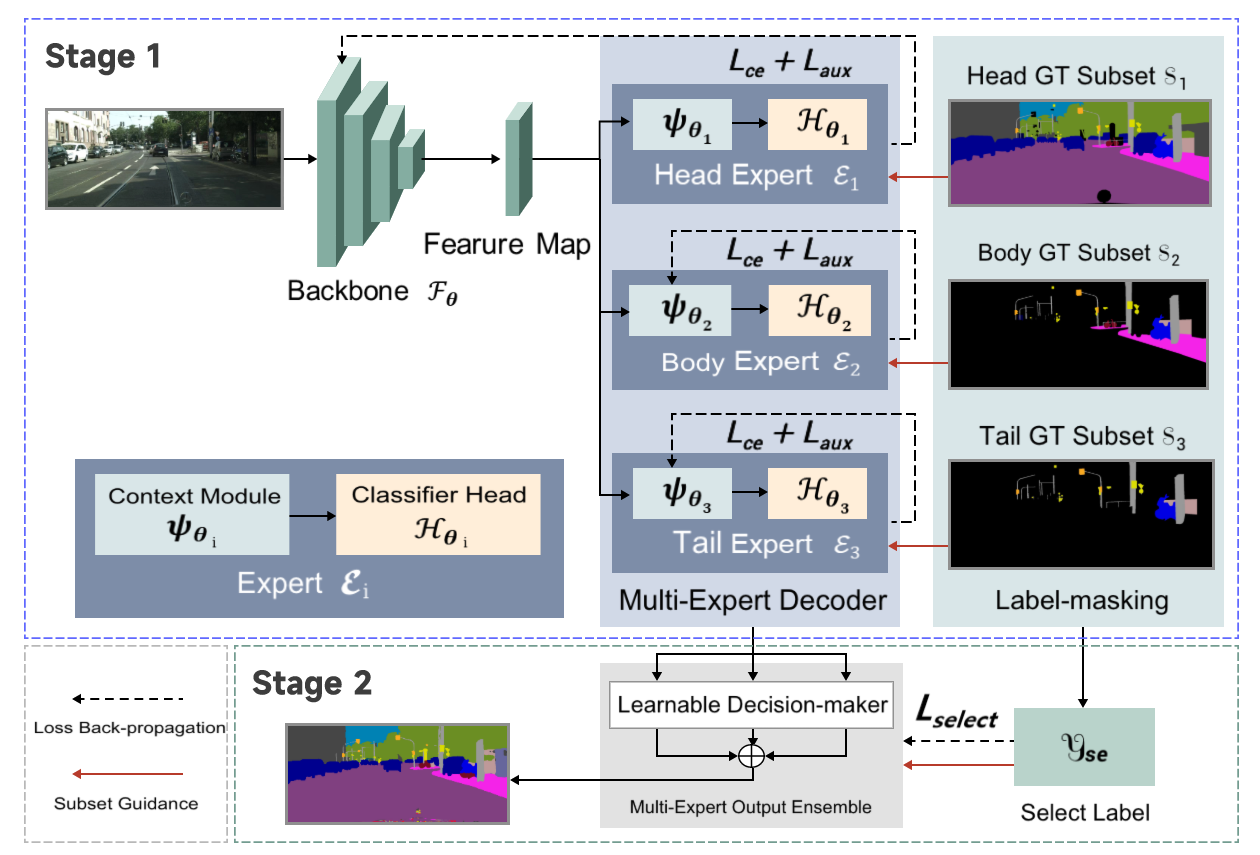}
    \vspace{-1em}
    \caption{The overview of the proposed framework. \textbf{Stage 1}, Followed by shared backbone for representation training, the prior distribution allocates a specific label-masking dataset for each expert with the prior knowledge, then a MED module combined with context module and classifier self-adaptive generates contextual information on the expert's dominated categories and classify pixels with $L_{ce}$ and $L_{aux}$. \textbf{Stage 2}, we generate the select label $y_{se}$ by prior distribution and MOE ensemble the outputs of all experts by a learnable decision-maker with the guidance of $y_{se}$. }
    \label{fig.framework}
    \vspace{-1em}
\end{figure*}

\section{RELATED WORK}
\label{gen_inst}

\subsection{Semantic Segmentation} 

FCN~\cite{r41-fcn} is regarded as the pioneer in the field. Because it introduces the full convolution on the whole image and formulates the semantic segmentation task as per-pixel classification with the basic framework of \emph{backbone \& context module}. Subsequently, various advanced methods have been introduced based on FCN, they can be roughly divided into two directions. One is to design a novel backbone for more robust feature representation~\cite{r42-hrnet,r43}. HRNet introduced a parallel backbone network to generate high-resolution representations. The other is to enrich contextual information for each pixel~\cite{r12-danet,r13,r14,r15-ocr}. For instance, combining high- and low-level features to extract global information~\cite{r44,r45-segnet}, introducing large receptive field, such as dilated or atrous convolutions~\cite{r11-deeplab,r16-pspnet} to gather multi-scale contextual cues and building the feature pyramids~\cite{r46-ppnet}. However, these approaches ignore the impact of data distribution, and our work is the first to explicitly focus on the long-tailed distribution in semantic segmentation.  
\subsection{Long-tailed Recognition} 

\paragraph{Re-Balancing} Re-weighting methods~\cite{r30,r31-oltr,r49-focal} adjust the loss function or boost larger weights on tail categories. While, re-sampling methods~\cite{r20,r33,r34} achieve data balance based on over-sampling the low-frequency categories, under-sampling the high-frequency categories, or even data augmentation, by generating additional samples to complement tail categories. However, both sample-wise and loss-wise methods increase the performance for the minor categories while compromising that for the major ones. 

\setlength{\parskip}{-0.25em}
\paragraph{Ensemble-and-Grouping} Recent studies indicate a trend of overcoming the long-tailed recognition problem using multi-expert and multi-branch strategies. BBN~\cite{r37-bbn} adopts two-branches to focus on normal and reversed sampling. LFME~\cite{r35-lfme}, RIDE~\cite{r24-ride}, and ACE~\cite{r48-ace} accomplish training on relatively balanced sub-groups. In these methods, the training data are divided by category frequency, and groups the classification results together into a multi-expert architecture. Those methods can learn diverse classifiers in a parallel way using knowledge distillation, distribution-aware expert selection, or complementary experts. Despite being suffered from re-balancing tricks, their success in image recognition has shown great potential for semantic segmentation.
\setlength{\parskip}{0em}

\section{Methods}\label{method}


As shown in Figure~\ref{fig_split}, each semantic category is tagged~\emph{head},~\emph{body}, or~\emph{tail}, according to its ranking regarding the frequency of the corresponding pixels. In particular, body categories refer to those ranked lower than head categories, but higher than tail categories. Figure~\ref{fig.framework} provides an overview of our two-stage MEDOE framework. Specifically, \textbf{Stage 1} adopts a backbone to extract a low-level feature map from the input image, whose contextual information is discovered by a multi-expert decoder (MED). The MED comprises three \emph{experts} corresponding to head, body, and tail categories (respectively referred to as \emph{head expert}, \emph{body expert}, and \emph{tail expert}). Each expert is composed of a context module and a classifier head. Implementing the proposed expert-specific label-masking strategy, each expert works on a unique dataset with the labels for certain categories masked, which ensures the expert focuses on its corresponding categories and predicts with higher confidence. 
We minimize both Cross-Entropy (CE) and auxiliary loss functions for each expert to constrain the effect of interfering pixels and the distribution of label-masking data. It is noteworthy that only the head expert updates the parameters of the backbone on the entire dataset, whereas the others refine their context modules;
\textbf{Stage 2} employs a multi-expert output ensemble (MOE) to aggregate the outputs from the experts for predicting the label of each pixel. The MOE uses a learnable decision-maker instead of user-specified weights.

\subsection{Stage 1: Multi-Expert Decoder} \label{MED}

\paragraph{Expert-Specific Label-Masking} For re-sampling that may undermine contextual information, we propose a novel pixel labeling strategy, which enables each expert to generate the contextual information from its dominated categories (\ie, head, body, or tail) in a self-adaptive way. The process can be defined as follows.

Given a training set $\mathbb{D} = \{(\mathit{X} , \mathit{Y}); \mathit{C}\}$, $\mathit{X}$ denotes the data, and $\mathit{Y}$ denotes ground truth labels with in total $\mathit{c}$ categories, where $\mathit{C} = \left\{1, 2, \ldots, \mathit{c} \right\}$. In particular, any category in $\mathit{C}$ is ascendingly sorted by the corresponding pixel frequency $F(\cdot)$ (\ie, $F(\mathit{i}) > F(\mathit{j}), \forall i > j$). For \textit{K} experts $\mathbb{E}=\left\{\mathcal{E}_{1}, \mathcal{E}_{2}, \ldots, \mathcal{E}_{K}\right\}$, we assign \textit{K} expert-specific datasets, where the labels for certain categories in each ground truth are masked. Their corresponding labeled categories $\mathbb{S}=\left\{\mathcal{S}_{1}, \mathcal{S}_{2}, \ldots, \mathcal{S}_{K}\right\}$. We set $\textit{K} = 3$, and $\mathcal{E}_{1}, \mathcal{E}_{2}$ and $\mathcal{E}_{3}$ dominate head, body and tail categories, respectively. Suppose $\mathit{c}_b$ and $ \mathit{c}_t$ denotes the first of body categories and that of tail categories, $\mathbb{S}$ can be further formulated as follows,
\begin{equation}
\setlength{\abovedisplayskip}{2pt}
\setlength{\belowdisplayskip}{3pt}
\begin{aligned}
    \mathcal{S}_{1} &= \left\{1, 2, \ldots, \mathit{c}\right\}, \\
    \mathcal{S}_{2} &= \left\{\mathit{c}_b, \mathit{c}_{b}+1, \ldots, c\right\}, \\
    \mathcal{S}_{3} &= \left\{\mathit{c}_t, \mathit{c}_{t}+1, \ldots, \mathit{c} \right\}.
    \label{eq:1}
\end{aligned}
\end{equation}

Our strategy exposes body and tail categories to their dominating experts. In addition, the overlapping strategy is also adopted to expose fewer categories to more experts. Therefore, the experts dominating the categories for difficultly segmentable pixels can integrate the others' outputs.

\setlength{\parskip}{-0.35em}
\paragraph{MED Architecture} Our MED aims to retrieve different contextual information from the feature map from the shared backbone ${f}_{\theta}$ to guide adjusting the boundaries of classifiers, following the observation that a model's performance for semantic segmentation tends to be superior for the majority of categories. In particular, the parameters of an expert are determined by its label-masking dataset, while the local optima for head categories can be avoided. Conventional context modules with large-scale receptive field convolutions $\Psi=\left\{\psi_{\theta_{1}}, \psi_{\theta_{2}}, \ldots, \psi_{\theta_{K}}\right\}$, such as PPM~\cite{r16-pspnet} and OCR~\cite{r15-ocr}, can self-adaptively generate different contextual information from dominant categories in the expert-specific label-masking datasets. Connected with classifier head $\mathbb{H}=\left\{\textit{h}_{\theta_{1}}, \textit{h}_{\theta_{2}}, \ldots, \textit{h}_{\theta_{K}}\right\}$, the expert $\mathcal{E}_{i}$ outputs prediction as follows,
\begin{equation}
    {z}_{i} = \textit{h}_{\theta_{i}}(\psi_{\theta_{i}}({f}_{\theta}(x))).
    \label{eq:2}
\end{equation}
\paragraph{Diverse Distribution-Aware Loss Function} As each expert is expected to achieve a satisfactory performance at least on its specific label-masking dataset, we adopt the Cross-Entropy (CE) loss function, which is aware of diversely distributed data instead of constraining the final prediction as in existent methods, to guide the per-pixel classification on the dataset $D_i$ corresponding to $\mathcal{S}_{i}$ for each expert $\mathcal{E}_{i}$ as follows,
\begin{equation}
\begin{gathered}
L_{ce}\left(z_{i},y_i;\theta_{i}\right)=-\textstyle{\sum}^{\mathcal{S}_{i}}_{c_j} y_{i}^{c_j} \log \left( \text{softmax}\left({z}_{i}^{c_j} 
\right)\right),\\
D_i = (x_i, y_i): \textstyle{\sum}_{n}^{N_{\mathcal{S}_i}} {x_{i_n}, y_{i_n}}, y_{i_n} \in \mathcal{S}_{i},
\end{gathered}
\label{eq:3}
\end{equation}

where $N_{\mathcal{S}_i}$ denotes the number of pixels' categories belong to $\mathcal{S}_{i}$. However, semantic segmentation adopts an end-to-end training pattern for each image, whereas some categories become interfering. Specifically, interfering categories $\mathcal{S}^{I C}_{i}$ for $\mathcal{E}_{i}$ refer to those categories excluded by $\mathcal{S}_{i}$ (i.e., $\mathcal{S}^{IC}_{i}\cup \mathcal{S}_{i} = \textit{C} \quad\& \quad\mathcal{S}^{IC}_{i}\cap \mathcal{S}_{i} = \emptyset $). Since interfering categories impede an expert from concentrating on its dominant categories, we devise an auxiliary loss function $L_{aux}$ with an L2 regularization term to suppress the impact of interfering categories while avoiding the over-confidence for dominant categories. Besides, we minimize the KL-divergence for each category ${c_j}\in \mathcal{S}_{i}$ between the classification probability $p_{i}^{c_j}$ and the probability $q^{c_j}$ in the ground truth dataset as follows,
\begin{equation}
\small
\begin{gathered}
L_{a u x}\left(z_{i},y_i;\theta_{i}\right) =\textstyle{\sum}_{c_{j} }^{\mathcal{S}^{I C}_{i}}\left\|{z}_{i}^{c_{j}}\right\| + \textstyle{\sum}_{c_{j}}^{\mathcal{S}_{i}} \mathit{p}_{i}^{c_{j}} \log \left(\frac{\mathit{p}_{i}^{c_{j}}}{\mathit{q}^{c_{j}}}\right), \\ 
\mathbf{{p}_{i}} = \text{softmax}({z}_{i}) \quad,\quad \mathbf{{p}_{i}} = [\mathit{p}_{i}^{c_{1}},\dots, \mathit{p}_{i}^{c_{j}},\dots].
\label{eq:4-2}
\end{gathered}
\end{equation}
Above all, the diverse distribution-aware loss function for each expert $\mathcal{E}_{i}$ is formulated as follows,
\begin{equation}
    L\left(z_{i},y_i;\theta_{i}\right) = L_{c e}\left(z_{i},y_i;\theta_{i}\right) + \alpha L_{a u x}\left(z_{i},y_i;\theta_{i}\right),
    \label{eq:5}
\end{equation}
where $\alpha$ is the hyper-parameter to balance $L_{ce}$ and $L_{aux}$, which is empirically set to $0.2$. Overall, the experts learn at Stage 1 are good and distinctive from each other, for their dominated categories are complementary.

\paragraph{Oracle Case} As mentioned, the MED, loss function, and expert-specific pixel-masking strategy enable each expert to effectively focus on their dominated categories. For the assumption of the Stage 1, the proposed framework achieves the theoretically optimal result when each pixel's label is predicted by the expert dominating its corresponding category, which is called the \emph{Oracle} case. This case is regarded as the theoretical upper bound of the proposed framework. Specifically, according to the overlapping strategy, when the ground truth label $\mathit{y}$ of pixel $\mathit{x}$ satisfies $\mathit{y} \in \mathcal{S}_{n}, \mathit{y} \notin \mathcal{S}_{n+1}$, the expert $\mathcal{E}_{n}$ is chosen to output the oracle classification probability $\mathbf{p}_{O}$ of instance $\mathit{x}$ as follows,
\begin{equation}
\setlength{\abovedisplayskip}{2pt}
    \mathbf{p}_{O} = \text{softmax}(\textit{h}_{\theta_{n}}(\psi_{\theta_{n}}({f}_{\theta}(x)))).
    \label{eq:5-2}
    \vspace{-1em}
\end{equation}

\begin{table*}[t]
\vspace{-0.5em}
    \centering
    \scalebox{1}{\begin{tabular}{l|c|cc|cc}
\hline \hline \multirow{2}{*}{Methods} & \multirow{2}{*}{Backbone} & \multicolumn{2}{c|}{Cityscapes} & \multicolumn{2}{c}{ADE20K} \\
 & &mIoU~(\%) & mAcc~(\%)& mIoU~(\%) & mAcc~(\%)\\
\hline DeepLabv3+~\cite{r11-deeplab} & ResNet-50c & $80.37$ & $86.68$ & $42.11$ & $54.13$\\
 DeepLabv3+$+$MEDOE  & ResNet-50c &$80.62\color{green}({+0.25})$ & $90.04\color{green}({+3.36})$  &$43.82\color{green}({+1.71})$ & $60.02\color{green}({+5.89})$\\
 DeepLabv3+$+$MEDOE$^{\dagger}$  & ResNet-50c &$84.14\color{green}({+3.77})$ &$92.38\color{green}({+5.70})$  & $53.34 \color{green}({+11.23}) $ & $73.97 \color{green}({+19.84})$\\
 PSPNet~\cite{r16-pspnet} & ResNet-50c &$78.32$ &$85.52$ &$40.46$ &$51.42$\\
 PSPNet$+$MEDOE & ResNet-50c &$78.82(\color{green}{+0.50})$ &$88.07(\color{green}{+2.55})$ &$41.76\color{green}({+1.30})$ &$54.27\color{green}({+2.85})$\\
 PSPNet$+$MEDOE$^{\dagger}$ & ResNet-50c &$84.32(\color{green}{+4.56})$ & $91.90(\color{green}{+5.34})$ &$52.00\color{green}({+11.54})$ & $71.61\color{green}({+20.19})$\\
 \hline
DeepLabv3+ & ResNet-101c & $80.67$ & $87.58$ & $44.60$ & $56.28$ \\
 DeepLabv3+$+$MEDOE & ResNet-101c & $81.21\color{green}({+0.54})$ & $91.29\color{green}({+3.69})$ & $46.13\color{green}({+1.42})$ & $61.12\color{green}({+4.84})$ \\
 DeepLabv3+$+$MEDOE$^{\dagger}$  & ResNet-101c &$84.51\color{green}({+3.84})$ &$92.63\color{green}({+5.05})$  &$55.18\color{green}({+10.58})$ &$76.82\color{green}({+20.54})$\\
 PSPNet & ResNet-101c &$79.76$ &$86.56$ &$43.33$ &$54.51$\\
 PSPNet$+$MEDOE & ResNet-101c &$79.79\color{green}({+0.03})$ &$90.88\color{green}({+4.32})$ &$44.31\color{green}({+0.98})$ &$59.85\color{green}({+5.19})$\\
 PSPNet$+$MEDOE$^{\dagger}$ & ResNet-101c &$84.32\color{green}({+4.56})$ & $91.90(\color{green}{+5.34})$ &$52.00(\color{green}{+11.54})$ & $71.61(\color{green}{+20.19})$\\
 \hline
OCRNet~\cite{r15-ocr} & HRNet-W48 & $80.70$ & $88.11$  & $42.53$ & $54.91$\\
 OCRNet$+$MEDOE & HRNet-W48 & $81.49\color{green}({+0.79})$ & $89.14\color{green}({+1.03})$ & $43.31\color{green}({+0.78})$ & $58.96\color{green}({+4.05})$ \\
 OCRNet$+$MEDOE$^{\dagger}$ & HRNet-W48 & $85.29\color{green}({+4.59})$ & $93.38\color{green}({+5.27})$ & $51.99\color{green}({+9.46})$ & $74.68\color{green}({+19.77})$\\
 \hline
 SegFormer~\cite{xie2021segformer} &MIT-B3 & $81.94$ &$88.28$ & $47.13$ &$60.84$\\
 SegFormer$+$MEDOE &MIT-B3 &$82.37(\color{green}{+0.43})$ &$91.16(\color{green}{+2.88})$ &$48.22\color{green}({+1.09})$ &$64.06\color{green}({+3.22})$\\
  SegFormer$+$MEDOE$^{\dagger}$ &MIT-B3 &$85.67\color{green}({+3.73})$ &$93.49\color{green}({+5.21})$ &$56.49\color{green}({+9.36})$ &$79.41\color{green}({+18.57})$ \\
\hline \hline
\end{tabular}
}
\vspace{-0.5em}
\caption{Comparison of performance on the validation set of Cityscapes and ADE20K with current methods. $\dagger$: Oracle results for the ideal case.}
\vspace{-1em}
\label{t2}
\end{table*}

\begin{table*}[t]
    \centering
    \scalebox{0.9}{\begin{tabular}{l|cc|cc}
\hline \hline \multirow{2}{*}{Methods} & \multicolumn{2}{c|}{Cityscapes} & \multicolumn{2}{c}{ADE20K}  \\
 &mIoU~(\%) &mAcc~(\%) &mIou~(\%) &mAcc~(\%) \\
\hline  Baseline (DeepLabv3+ \& ResNet-50c) & $80.37$ & $86.68$ & $42.11$ & $54.13$\\
Baseline$+$Re-weighting(Focal Loss)~\cite{r49-focal}  &$76.23\color{red}({-4.14})$ & $85.00\color{red}({-1.68})$ & $37.61\color{red}({-4.50})$ & $55.29\color{green}({+1.16})$ \\
Baseline$+$Re-weighting(LDAM Loss)~\cite{LDAM}  &$77.52\color{red}({-2.85})$ &$85.15\color{red}({-1.43})$ &$40.39\color{red}({-1.72})$ &$48.78\color{red}({-5.35})$\\
Baseline$+$Re-weighting(Seesaw Loss)~\cite{wang2021seesaw}  &$67.58\color{red}({-12.79})$ &$74.35\color{red}({-12.33})$ &$33.87\color{red}({-8.24})$ &$40.19\color{red}({-13.94})$\\
Baseline$+$Re-sampling(Under sampling) &$66.79\color{red}({-13.58})$ &$75.21\color{red}({-11.47})$ & $23.12\color{red}({-18.99})$ & $25.11\color{red}({-29.02})$ \\
Baseline$+$MEDOE &$\mathbf{80.62\color{green}({+0.25})}$ & $\mathbf{90.04\color{green}({+3.36})}$ &$\mathbf{43.82\color{green}({+1.71})}$ &$\mathbf{60.02\color{green}({+5.89})}$\\
\hline \hline
\end{tabular}}
\caption{Comparison of performance between our method with pixel-level re-weighting and re-sampling methods. Results in \textbf{bold} denote the best performance. }
\vspace{-1em}
\label{t8}
\end{table*}

\begin{figure*}[t]
\label{viso_ade1}
    \centering
    \subfigure[Image]{
    \begin{minipage}[t]{0.22\linewidth}
    \centering
    \includegraphics[width = 1.4in,height = 2in]{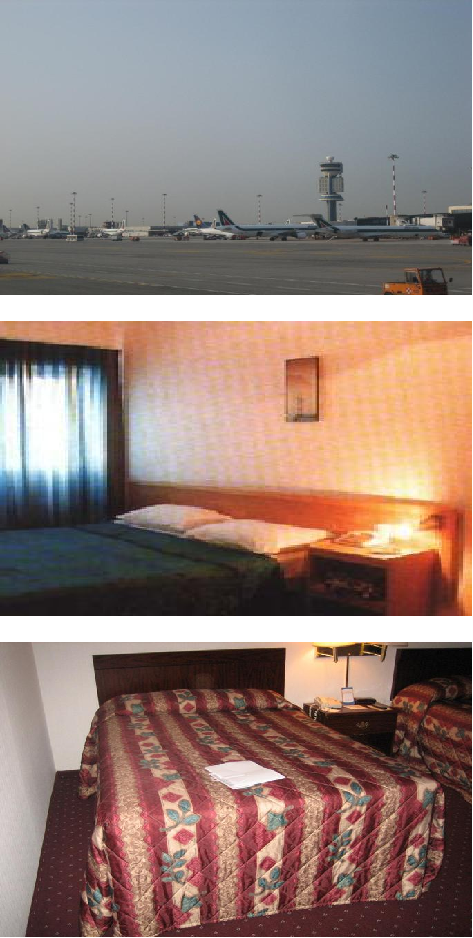}
    \end{minipage}}
    \subfigure[Ground Truth]{
    \begin{minipage}[t]{0.22\linewidth}
    \centering
    \includegraphics[width = 1.4in,height = 2in]{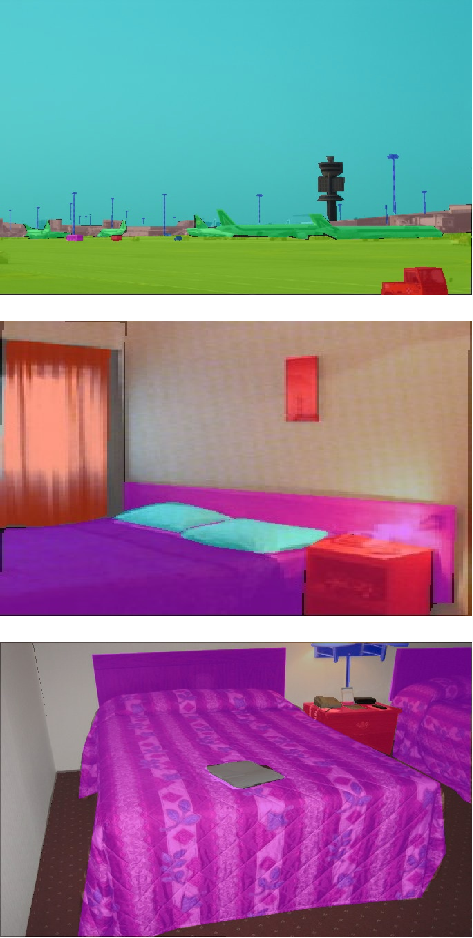}
    \end{minipage}}
    \subfigure[DeepLabv3+]{
    \begin{minipage}[t]{0.22\linewidth}
    \centering
    \includegraphics[width = 1.4in,height = 2in]{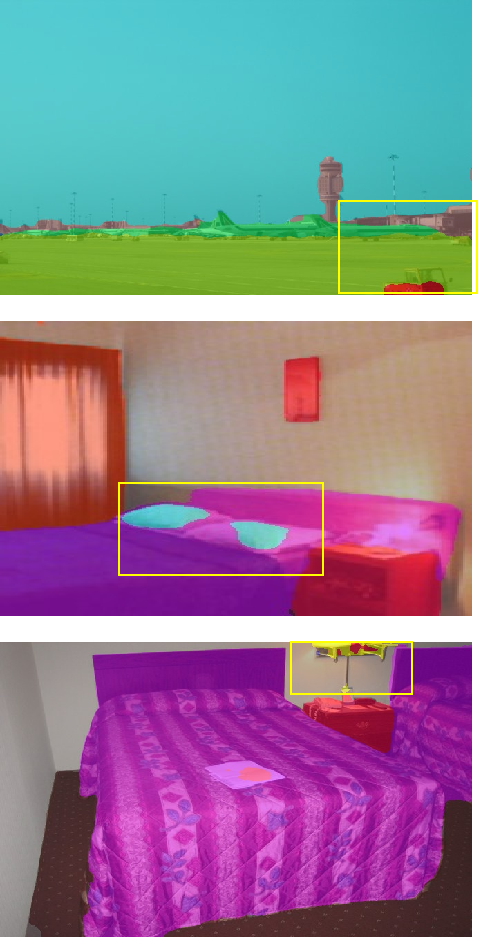}
    \end{minipage}}
    \subfigure[MEDOE~(Ours)]{
    \begin{minipage}[t]{0.22\linewidth}
    \centering
    \includegraphics[width = 1.4in,height = 2in]{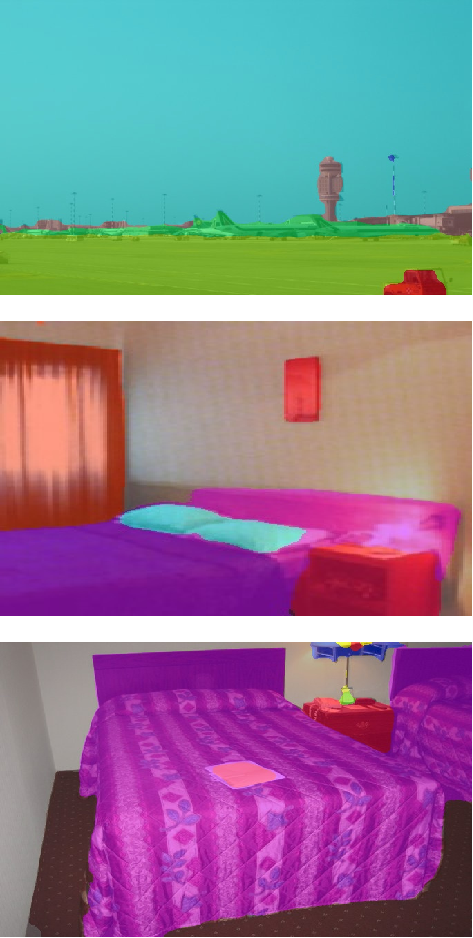}
    \end{minipage}}
    \vspace{-1em}
    \caption{Qualitative Visualization on the validation set of ADE20K with ResNet-50c as backbone. All the models here are trained under the same setting. }
    \label{fig.ade_vis}
    \vspace{-1em}
\end{figure*}

\begin{table*}[t]
\setlength\tabcolsep{2.5pt}
    \centering
    \scalebox{1}{\begin{tabular}{l|l|cc|cc|cc|cc}
        \hline \hline 
        \multirow{2}{*}{Dataset} &
        \multirow{2}{*}{Methods} & \multicolumn{2}{c|}{Overall} &
        \multicolumn{2}{c|}{Head} & \multicolumn{2}{c|}{Body} &
        \multicolumn{2}{c}{Tail}\\
        & &mIoU(\%) & mAcc(\%) &mIoU(\%) & mAcc(\%) &mIoU(\%) & mAcc(\%) &mIoU(\%) & mAcc(\%)\\
\hline \multirow{4}{*}{Cityscapes} & Baseline & ${80.37}$ & $86.68$ &$\textbf{95.26}$ &$\textbf{97.74}$ & ${83.42}$ & $91.49$ & $73.01$ & $81.43$ \\
& Baseline$+$Focal Loss &$76.23$ &$85.00$ &$91.18$ &$90.97$ &$80.55$ & $90.33$ & $68.79$ &$84.56$\\
& Baseline$+$MEDOE &$\textbf{80.62}$ & $\textbf{90.04}$  & $94.87$ &$97.54$ &$\textbf{83.74}$ &$\textbf{91.72}$ &$\textbf{74.30}$ &$\textbf{87.20}$\\
& Baseline$+$MEDOE$^{\dagger}$ &$\textit{84.14}$ &$\textit{92.38}$ & $\textit{96.43}$ &$\textit{97.72}$ & $\textit{87.12}$ &$\textit{96.70}$ & $\textit{78.59}$ &$\textit{88.88}$\\
\hline
\multirow{4}{*}{ADE20K} & Baseline& $42.11$ & $54.13$ & $65.50$ & $\textbf{78.71}$ & $44.75$ &$59.75$ & $33.64$ &$42.97$   \\
& Baseline$+$Focal Loss &$37.61$ &$55.29$ &$60.01$ &$74.89$ &$41.96$ &$61.28$ &$30.28$ &$46.95$\\
& Baseline$+$MEDOE &$\textbf{43.82}$ & $\textbf{60.02}$ & $\textbf{66.99}$ &$77.90$ &$\textbf{46.32}$ &$\textbf{63.52}$ &$\textbf{35.38}$ &$\textbf{52.33}$  \\
& Baseline$+$MEDOE$^{\dagger}$& $\textit{53.34} $
 & $\textit{73.97} $  &$\textit{67.38}$ &$\textit{75.49}$ & $\textit{55.65}$ &$\textit{74.84}$ & $\textit{46.75}$ &$\textit{74.09}$\\
\hline \hline
\end{tabular}
}
\caption{Comparison of performance among different categories data subsets from Cityscapes and ADE20K with DeepLabv3+ \& ResNet-50c. \(\dagger\): Oracle results for the ideal case. Results in \textbf{bold} denote the best performance. \label{t_orcal}}
\vspace{-1.5em}
\end{table*}
\setlength{\parskip}{0em}
\subsection{Stage 2: Multi-Expert Output Ensemble}
MOE is designed upon the unavailability of oracle information on decisions and aims to aggregate the classification probabilities. Specifically, our MOE takes as input the outputs from Stage 1 and makes a decision on the ensemble of predictions. In the ideal case, an expert $\mathcal{E}_{i}$ should be selected if the ground truth label $y$ of the current pixel $x$ belongs to $\mathcal{E}_{i}$'s dominated categories. However, training for such a decision on expert selection is difficult due to it heavily relies on data distribution. Instead, we design a category-specific linear classifier for the decision-maker. In particular, classification probabilities $\mathbf{p}_{i} = [\mathit{p}_{i}^{1}, \mathit{p}_{i}^{2}, \dots, \mathit{p}_{i}^{c}]$ are adjusted for $\mathit{x}$ with $\mathcal{E}_{i}$ as follows,
\begin{equation}
    \hat{p}_{i}^{j} = \text{softmax}(\mathit{w}_{i}^{j} \mathit{p}_{i}^{j} + \mathit{\beta}_{i}^{j}),
\end{equation}
where $\mathit{w}^{j}$ and $\mathit{\beta}^{j}$ are calibration parameters for each category to be learned from data, $\mathit{w}_{i}^{j}$ denotes the selected decision weight for $\mathcal{E}_{i}$ on category $\mathit{j}$.

We introduce a penalty item $\mathit{\beta}^{j}$ to correct the over-confidence for tail categories. The decision-maker is optimized with the CE loss in training to minimize the ensemble prediction $\mathbf{p}_{final}$ in the Kronecker delta value $\mathit{y}_{se}$, which equals 1 if $\textit{y}_{i}$ = \textit{c} or 0 otherwise, as follows, 
\begin{equation}
\begin{aligned}
\mathbf{p}_{final} &= \frac{1}{K}\textstyle{\sum}^{\textit{K}}_{i=1}\mathbf{w}_{i}\mathbf{\hat{{p}_{i}}} \\
L_{select}\left(\mathbf{p}_{final},\mathit{y}_{se}\right)&=-\textstyle{\sum}^{ \textit{N}}_{n=1} \mathit{y}_{se}^{n} \log \left(\mathbf{{p}}^{n}_{final}\right).
\end{aligned}
\label{eq:5-4}
\end{equation}

\section{Experiments}
\label{exp}

To evaluate the effectiveness of the proposed framework, we conducted the experiments integrating our MEDOE and other long-tailed recognition methods with DeepLabv3+~\cite{r11-deeplab}, PSPNet~\cite{r16-pspnet}, OCRNet~\cite{r15-ocr} and SegFormer~\cite{xie2021segformer} on two challenging datasets: Cityscapes~\cite{r38-2016cityscapes} and ADE20K~\cite{r39-ade20k}. Due to the limit, please  refer to the Appendix as supplementary material for details on benchmarks $\&$ implementation.

\subsection{Comparison with State-of-the-Arts} \label{dataset}
 \textbf{Cityscapes}~\cite{r38-2016cityscapes} is a large-scale dataset containing a diverse set of urban street scenes with a long-tailed distribution due to certain rare categories (\eg, poles and trucks). As shown in Table~\ref{t2} and~\ref{t_orcal}, integrating existent methods with the proposed framework significantly improved the performance for semantic segmentation in mAcc and mIoU, compared to only using themselves. Specifically, integration with our MEDOE achieved an increase by up to 0.54\%, 0.50\%, 0.79\%, and 0.43\% in mIoU, and by up to 3.69\%, 4.32\%, 1.03\% and 2.88\% in mAcc for DeepLabv3+, PSPNet, OCRNet, and SegFormer, respectively.

 \begin{table*}[t]
\setlength\tabcolsep{2.5pt}
    \centering
    \scalebox{1}{\begin{tabular}{l|l|cc|cc|cc|cc}
        \hline \hline 
        \multirow{2}{*}{Datasets} &
        \multirow{2}{*}{Methods} & \multicolumn{2}{c|}{All} &
        \multicolumn{2}{c|}{Head} & \multicolumn{2}{c|}{Body} &
        \multicolumn{2}{c}{Few}\\
         & & mIoU(\%) & mAcc(\%) & mIoU(\%) & mAcc(\%) & mIoU(\%) & mAcc(\%) & mIoU(\%) & mAcc(\%)\\
         \hline 
        \hline \multirow{2}{*}{Cityscapes-uniform} & Baseline & $80.42$ & $86.98$ & $95.01$ & $\textbf{97.74}$ & $83.61$ &$91.48$ & $73.96$ &$81.43$ \\ 
         & +MEDOE & $\textbf{81.93}$ & $\textbf{90.86}$ & $\textbf{95.31}$ &$97.46$ &$\textbf{84.34}$ &$\textbf{92.75}$ &$\textbf{76.18}$ &$\textbf{87.74}$ \\
         \hline
        \multirow{2}{*}{ADE20K-uniform} & Baseline & $44.25$ & $55.19$ & $57.12$ & $\textbf{79.76}$ & $44.97$ &$59.87$ & $40.01$ &$44.75$ \\
         & +MEDOE &$\textbf{48.44}$ & $\textbf{60.30}$ & $\textbf{58.83}$ &$78.44$ &$\textbf{46.54}$ &$\textbf{61.27}$ &$\textbf{46.90}$ &$\textbf{54.42}$ \\
        \hline
    \end{tabular}}
    \caption{Comparison MEDOE with baseline (DeepLabv3+ \& ResNet-50c) in the uniform distribution. Results in \textbf{bold} denote the best performance.}
    \vspace{-1em}
    \label{t_uniform}
\end{table*}

\begin{table*}[t]
\setlength\tabcolsep{2.5pt}
    \begin{minipage}{0.65\textwidth}
    \scalebox{0.9}{\begin{tabular}{l|cc|cc|cc|cc|c}
        \hline \hline 
        \multirow{2}{*}{Method} & \multicolumn{2}{c|}{All} &
        \multicolumn{2}{c|}{Head} & \multicolumn{2}{c|}{Body} &
        \multicolumn{2}{c|}{Few}&
        \multirow{2}{*}{Params}\\
        & mIoU &mAcc & mIoU &mAcc & mIoU &mAcc & mIoU &mAcc\\
        \hline Baseline & $42.11$ & $54.13$ & $65.50$ & $\textbf{78.71}$ & $44.75$ &$59.75$ & $33.64$ &$42.97$ &$25.82(1.0\times)$\\ 
        Baseline$+$MCN & $40.15$ & $53.79$ & $62.01$ &$75.51$ &$42.53$ &$56.20$ &$32.15$ &$45.18$ &$28.39(1.1\times)$\\
        Baseline$+$ME & $42.50$ & $54.46$ & $65.59$ & $78.62$ & $45.00$ &$60.31$ & $33.87$ &$42.95$ &$77.98(3.0\times)$\\
        Baseline$+$MED &$\textbf{43.82}$ & $\textbf{60.02}$ & $\textbf{66.99}$ &$77.90$ &$\textbf{46.32}$ &$\textbf{63.52}$ &$\textbf{35.38}$ &$\textbf{52.33}$ &$35.49(1.4\times)$\\
        \hline \hline
    \end{tabular}}
    \vspace{-0.5em}
    \captionof{table}{Ablation of \textbf{MCN}, \textbf{ME} method and \textbf{MED} with DeepLabv3+ \& ResNet-50c on ADE20K to verify the effectiveness of \textit{multi-context decoder} module. Results are reported in mIoU~(\%), mAcc~(\%), and Params~(GBs).}
    \label{t3}
    \vspace{-1em}
    \hspace{1em}
\end{minipage}
\hspace{1em}
\setlength\tabcolsep{2.5pt}
\begin{minipage}{0.30\textwidth}
    \vspace{0.0em}
    \centering
       \scalebox{0.9}{\begin{tabular}{l|c|c}
        \hline \hline \multirow{2}{*}{Aggregation} &\multirow{2}{*}{mIoU(\%)} & \multirow{2}{*}{mAcc~(\%)}  \\
        & &\\
        \hline 
        MOE &$\textbf{80.62}$ &$\textbf{90.04}$ \\
        Softmax(1) & $72.31$ & $82.32$ \\
        Argmax(2) &$70.45$ & $83.14$  \\
        Group Avg(3) &$78.87$ &$86.49$\\
        \hline \hline
    \end{tabular}}
    \vspace{-0.5em}
    \captionof{table}{Ablation study on outputs ensemble methods with DeepLabv3+ \& ResNet-50c on Cityscapes. }
    \label{t6}
\end{minipage}
\vspace{-1.5em}
\end{table*}
\setlength{\parskip}{-0.5em}
\paragraph{ADE20K}~\cite{r39-ade20k} is a challenging dataset containing scene-centric images at various scales and of diverse categories, causing a long-tailed distribution. As shown in Table~\ref{t2} and~\ref{t_orcal}, integrating other methods with our MEDOE also improved the performance similar to the case on Cityscapes. Specifically, integration with the proposed framework achieved an increase by up to 1.71\%, 1.30\%, 0.78\%, and 1.09\% in mIoU, and by up to 5.89\%, 5.19\%, 4.05\% and 3.22\% in mAcc for DeepLabv3+, PSPNet, OCRNet, and SegFormer, respectively. In addition, the example outputs on the validation set of ADE20K are illustrated in Figure~\ref{fig.ade_vis}, demonstrating the effectiveness of the proposed framework in accomplishing the segmentation with higher quality, especially for the tail categories (\eg, ``pillows" and ``lamps") in different scenes.

\paragraph{Comparison with Re-Balancing Methods} As shown in Table~\ref{t8}, it is not feasible to directly apply the current pixel-level re-balancing methods in long-tailed semantic segmentation. Specifically, we implemented the re-weighting method by modifying the loss function and assigning larger weights to the tail categories. The re-weighting methods cause significant decreases in mIoU, especially in the performance for head categories. For re-sampling, the contextual information is corrupted resulting in segmentation metrics at a very low level. In general, only our MEDOE framework achieved impressive results.

\begin{figure}[t]
\label{improvements}
    \centering
    \subfigure[Performance improvement on Cityscapes]{
    \centering
    \includegraphics[height = 1.2in, width = 2.8in]{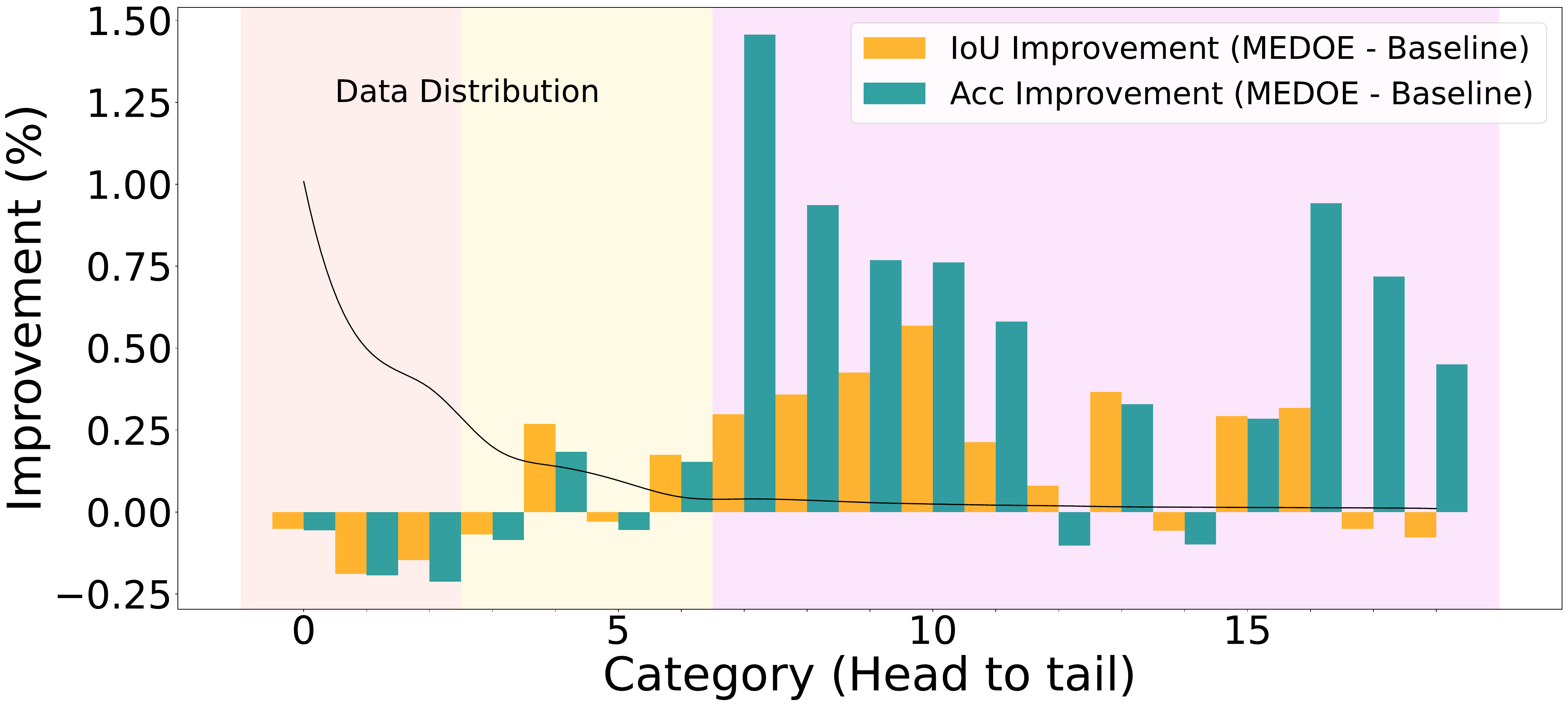}
    }
    \subfigure[Performance improvement on ADE20K]{
    \centering
    \includegraphics[height = 1.2in, width = 2.8in]{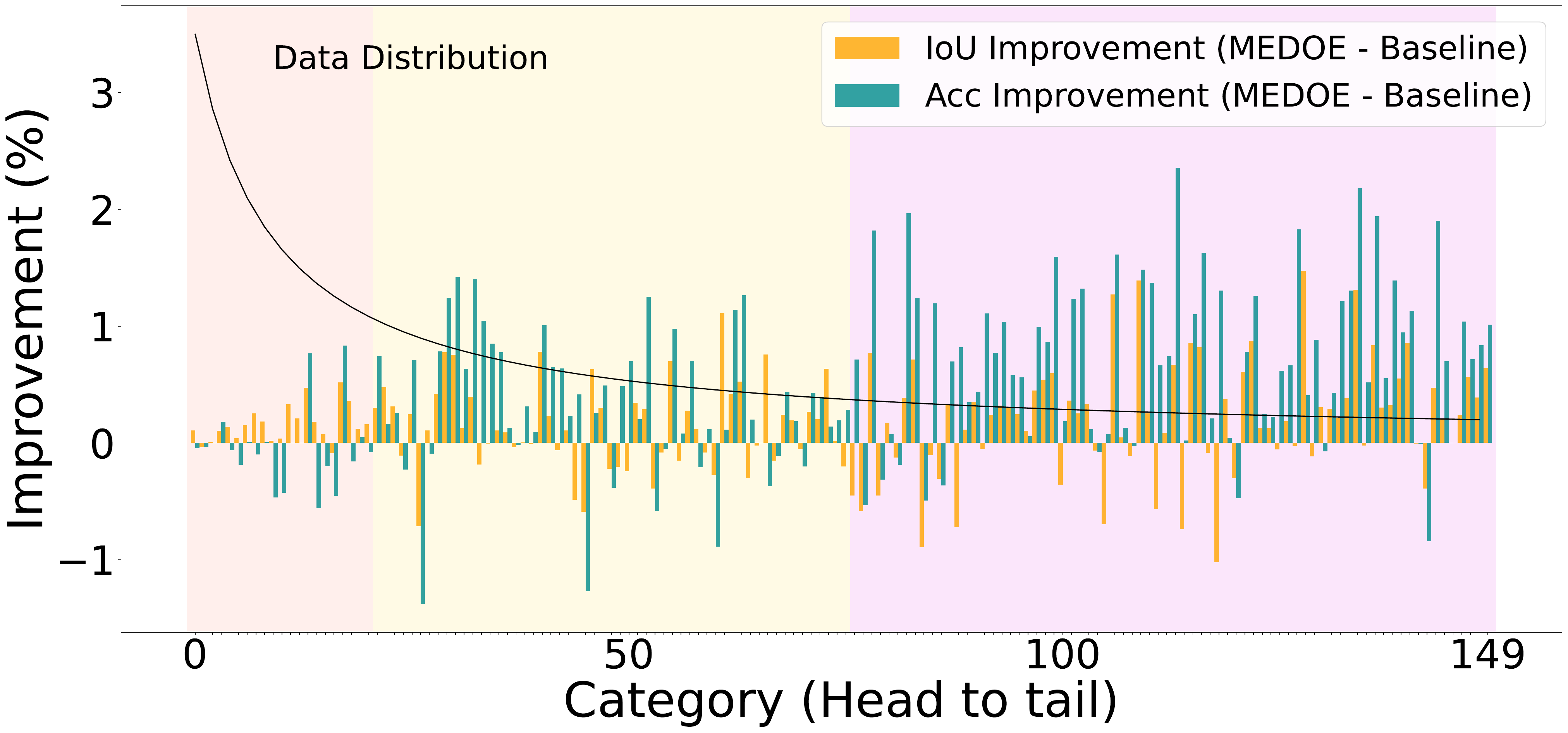}
    }
    \vspace{-1em}
    \caption{Comparisons of each category performance on Cityscapes and ADE20K with MEDOE and baseline. Our MEDOE gains both IoU and Acc improvements in body and tail categories, as well as avoids decreasing the head performance. }
    \label{fig.metric}
    \vspace{-2em}
\end{figure}
    

\paragraph{Comparison among Different Categories} As shown in Table~\ref{t_orcal}, our method achieved impressive increases in mIoU and mAcc for body and tail categories on Cityscapes and ADE20K, especially on the latter. In addition, Figure~\ref{fig.metric} illustrated the comparison of category-wise performance, and our MEDOE outperformed the baseline on both datasets for almost all categories. In particular, the advantage of the proposed framework is more significant for body and tail categories, verifying the MED's capability of extracting effective contextual information for per-pixel classification. 

\paragraph{Results in Oracle Case} As shown in Table~\ref{t2} and~\ref{t_orcal}, the proposed framework reached more impressive and consistent improvement in mIoU and mAcc on both datasets in the \textit{Oracle} Case. Specifically, our MEDOE brought an increase by 3\%-4\% in mIoU and 5\%-7\% in mAcc on Cityscapes, and by 11\%-13\% in mIoU and 19\%-22\% in mAcc on ADE20K. This considerable improvement suggested that the experts trained at Stage 1 had managed to focus on their dominated categories. Meanwhile, adopting the expert-specific label-masking strategy in the inference phase could largely outperform state-of-the-art methods, which shown that the MED is promising and more efforts could be explored to constrain the experts' decisions.

\paragraph{Extension with Long-tailed Recognition Setting} \label{uniform_set}
As mentioned in the introduction, we followed the long-tailed recognition setting that requires the training set to be in a long-tailed distribution and the test set to be in a uniform distribution. In particular, we sampled the long-tailed test set into a uniform distribution using the label-masking strategy. As shown in Table~\ref{t_uniform}, our MEDOE achieved a more significant improvement under the uniform distribution, especially for the tail categories, with a less considerable compromise on the accuracy for the head categories. It suggested that MEDOE is fairer to the tail and body categories and is robust to the change in data distribution by balancing the contribution for each category, instead of focusing on specific distributions. Overall, our MEDOE can be applied to extensive data distributions.
\subsection{Ablation Studies \label{ablation}}
\setlength{\parskip}{-0.25em}
\paragraph{Effectiveness of MED} To verify that our MED is rather than modifying the classifiers' boundaries or adopting model-ensemble methods, we integrated the baseline with MED, a multi-classifier network \textbf{(MCN)} with a shared backbone and context module, and a model-ensemble \textbf{(ME)} method for comparison on Cityscapes. As shown in Table~\ref{t3}, only modifying the classifiers' boundaries using the MCN is not effective for tail categories and even reduces the overall performance. The ME method ignores tail categories in the segmentation task, although slightly improving the performance both overall and for head categories because of numerous parameters in multiple backbones. In comparison, our MED enables the baseline to better classify the pixels using different contextual information, which is adaptively extracted under the expert-specific label-masking strategy.

\paragraph{Effectiveness of MOE} We compared our MOE with several user-specified ensemble methods in Table~\ref{t6}. For each pixel $\mathit{x}$, (1) the Softmax method outputs ${z}_{i}$ for $\mathcal{E}_{i}$ and pixel category ${c}_{i}=j$ if $\text{softmax}({z}_{i}^{j}) > \beta$, $\beta$ is set as 0.3; (2) Argmax method determines the outputs by the confidence of the maximum softmax on each expert; (3) Group average method ensembles \({z}_{i}\) of $\mathcal{E}_{i}$ by grouping average weights. Generally, MOE achieved improvements overall. 

\begin{figure}
    \centering
    \includegraphics[height = 1.3in, width = 2.8in]{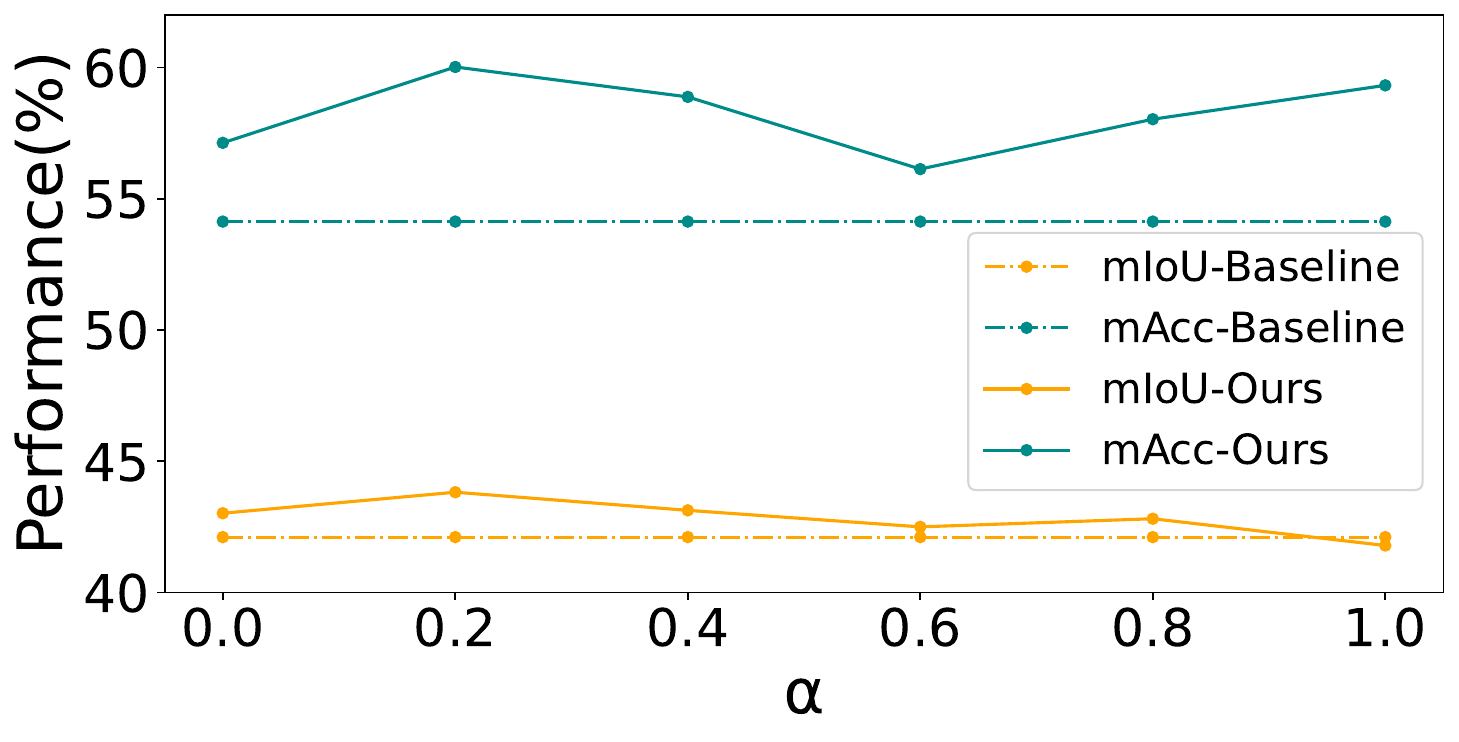}
    \vspace{-1em}
    \caption{Ablation w/ or w/o Auxiliary Loss and the hyper-parameter $\alpha$ on ADE20K with DeepLabv3+ \& ResNet-50c.}
    \label{fig:loss}
    \vspace{-1em}
\end{figure}

\paragraph{Effectiveness of Auxiliary Loss Function} We conducted an ablation with the auxiliary loss function on ADE20K to verify its effectiveness. As illustrated in Figure~\ref{fig:loss}, the improvement by our MEDOE still existed but is limited when solely using the CE loss function (\ie, $\alpha$=0). Meanwhile, our MEDOE attained its best performance when $\alpha=0.2$. We interpreted this observation as that the model trained without the auxiliary loss function can be more likely confused by interfering categories and thus degenerate regarding the performance overall. Therefore, introducing the auxiliary loss function may contribute to a more consistent and stable improvement in performance.

\setlength{\parskip}{0em}

\section{Conclusion}
In this paper, we investigated and identified the long-tailed distribution in semantic segmentation, motivated by the fact that existing methods ignore this issue and the performance declines for tail categories. We proposed a MEDOE framework to overcome this challenging problem. The MED, MOE module, and a series of strategies (\eg, expert-specific pixel-masking strategy and diversely distributed data-aware loss function.) were introduced to help each expert extract different contextual information and promote the performance in dominant categories. Our method achieved great improvements across different current methods up to 1.78\% gains in mIoU and 5.89\% gains in mAcc on Cityscapes and ADE20K. Furthermore, training a robust output ensemble decision-maker appears to be a very promising way of constraining the inference, as demonstrated with the significantly improved results obtained by the oracle. 

\newpage
\bibliography{aaai24}

\clearpage
\appendix
\section{Appendix \label{appendix}}

\subsection{Benchmarks \& Implementation Details} \label{bench}
\setlength{\parskip}{-0.25em}
\subsubsection{Benchmarks:} \paragraph{Cityscapes}~\cite{r38-2016cityscapes} is a dataset that focuses on the semantic understanding of urban street scenes and contains 19 semantic categories. It contains 5K annotated images with pixel-level fine annotations and 20K coarsely annotated images. The finely annotated 5K images are split into sets with numbers 2975, 500, and 1525 for training, validation, and testing. 

\paragraph{ADE20K}~\cite{r39-ade20k} is a challenging scene parsing dataset. It contains 150 categories and diverse scenes with 1038 image-level labels. The training and validation sets consist of 20K and 2K images, respectively.

\subsection{Implementation Details} We initialize backbones with the weights pre-trained on ImageNet. For DeepLabv3+, we using CNN-based ResNet-50c and ResNet-101c as backbones, which switch the first 7\(\times\)7 convolution layer to three 3\(\times\)3 convolutions. HRNet-W48~\cite{r42-hrnet} is adopted for OCRNet. MIT-B3 as a Transformer-based backbone is adopted for SegFormer~\cite{xie2021segformer}, both of which are popular in semantic segmentation. For CNN-based network, we use SDG and poly learning rate schedule~\cite{r16-pspnet} with factor $\left(1-\frac{ { iter }}{{ total_{iter} }}\right)^{0.9}$. The initial learning rate is set as 0.01 and the weight decay is 0.0005. We adopt AdamW with $6\times10^{-5}$ learning rate and 0.01 weight decay. We set the image crop size to 512\(\times\)1024, batch size as 8, and training iterations as 80K on Cityscapes by default. For ADE20K, the crop size of images is set as 512\(\times\)512, the batch size is set as 16, and training iterations are set as 80K if not stated otherwise. Especially, at Stage 2, we set 1K6 iterations for training MOE modules. In the whole training phase, we augment data samples with the standard random scale in the range of [0.5, 2.0], random horizontal flipping, random cropping, as well as random color jittering. For inference, the input image size of ADE20K is the same as the size during training, but for Cityscapes the input image is scaled to 1024\(\times\)2048, no tricks(\eg multi-scale with flipping) will be adopted during testing. All experiments are implemented on the Nvidia A6000.

\paragraph{Experiments details}
In this section we state some details of the experiments setting.
\paragraph{Expert-specific pixel-masking strategy details} We calculated the distribution of pixels by category in the whole dataset and grouped the pixels by the criterion of data magnitude. \ie, in ADE20K, head categories refer to those $\geq 1\%$, body categories to those $\in [0.1\%, 1\%)$, and tail categories to those $< 0.1\% $.
\paragraph{Details about ablation of MED} We adopted multi-classifier network (MCN) and model-ensemble (ME) method for comparison. In detail, ME is a model ensemble with $3$ same segmentation networks while averaging the outputs, and MCN is a shared backbone and context module, followed by three not shared classifiers for classifying head, body and tail, respectively.

\subsection{Additional Proof of the Correlation between Performance and Data Distribution}

\begin{figure}[th]
    \centering
    \subfigure[Data distribution and performance with the baseline on Cityscapes. \label{fig_a}]{
    \centering
    \includegraphics[width = 2.6in]{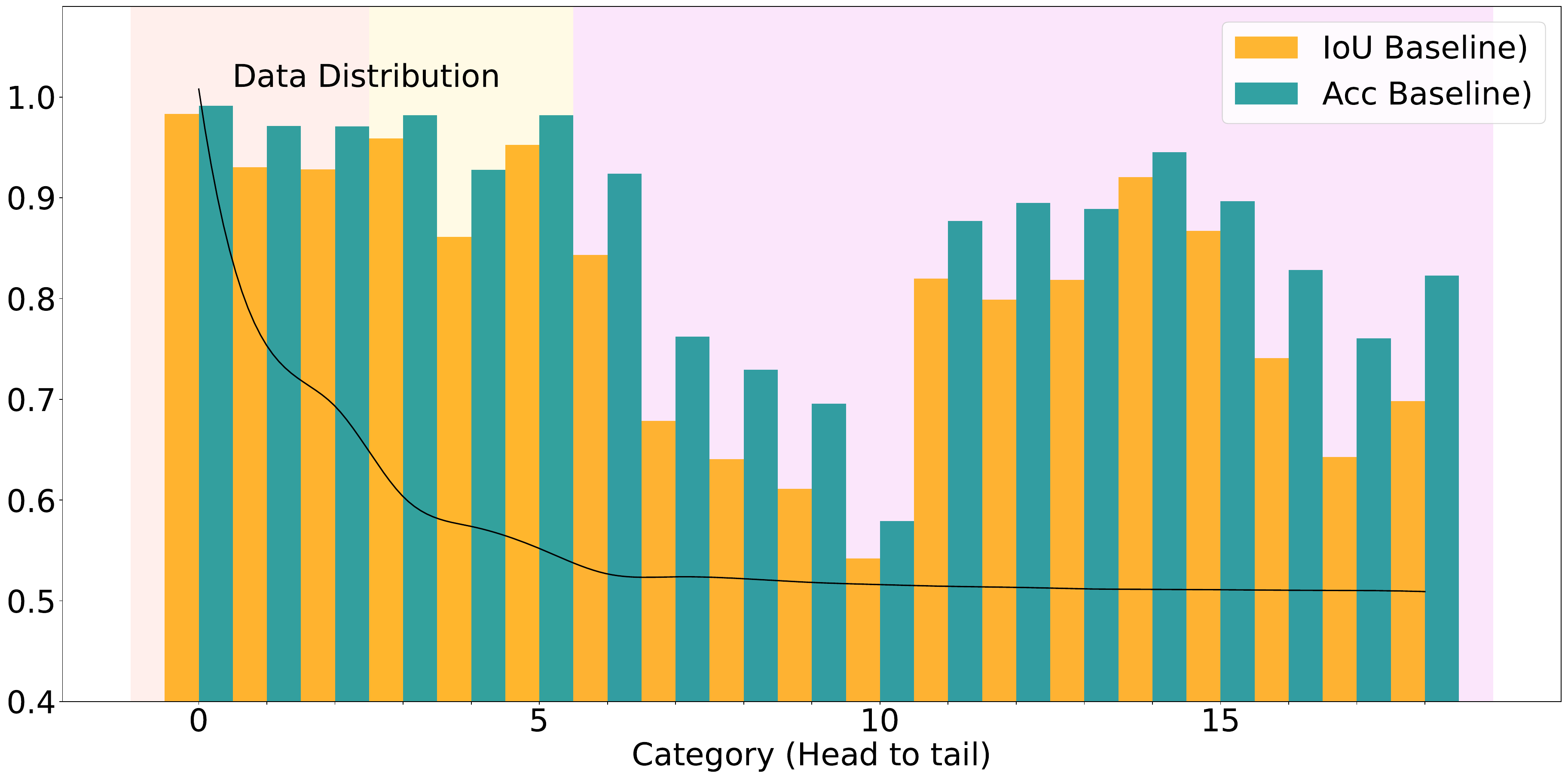}
    }
    \subfigure[Data distribution and performance with baseline on ADE20K. \label{fig_b}]{
    \centering
    \includegraphics[width = 2.6in]{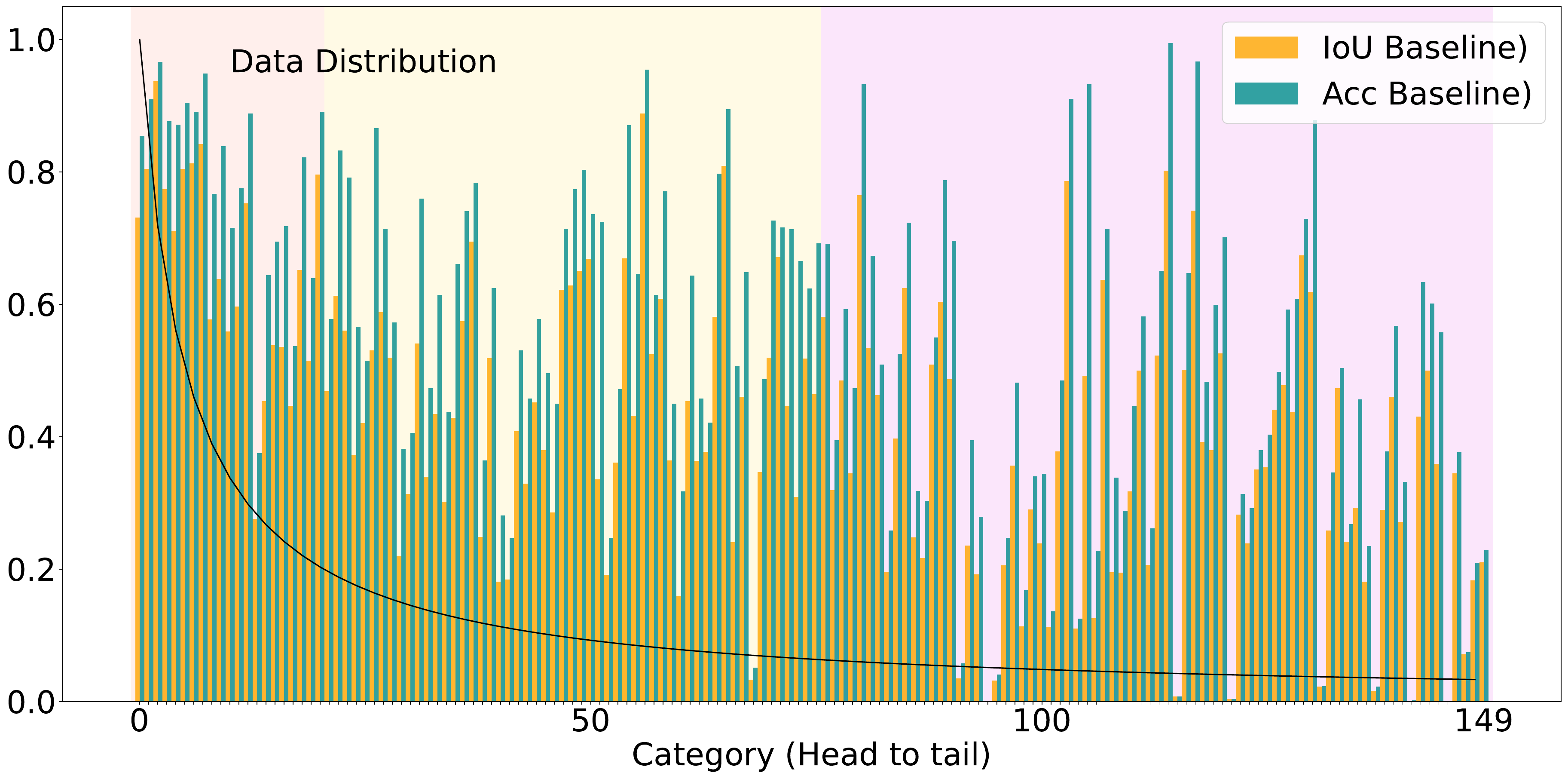}
    }
    \caption{The existence of long-tailed distribution in semantic segmentation, we set the baseline as DeepLabv3+ \& ResNet-50c by default. (a) and (b) show the data distributions on Cityscapes and ADE20K are long-tailed, which causes better performance on the majority categories yet suppresses the minority categories. It should be noted that we reordered the categories in Cityscapes according to Pixel Frequency.\label{fig_category_acc} } 
    \vspace{-1em}
\end{figure}

As shown in Figure~\ref{fig_category_acc}, the baseline method performs not well on certain categories on Cityscapes and ADE20K, and we can clearly see that these categories mainly fall in the tail and body data subsets. This further demonstrates that the long-tailed data distribution limits the overall performance of the baseline method by constraining the accuracy of certain categories. 

\subsection{Ablation of overlapping strategy}

We not only theoretically analyzed the benefits of choosing an overlapping strategy. We provided an ablation study with DeepLabv3+ $\&$ ResNet-50c on ADE20K in Table~\ref{overlap}, showing the superiority of the overlapping strategy.

\subsection{Ablation with State-of-the-Art Methods in Parameters}

In this section, we further show a comparison of our MEDOE with previous methods in Tabel~\ref{paramt1} and~\ref{paramt2}. Combined with Table~\ref{t8} in Sec~\ref{dataset}, we demonstrate that simply increasing the number of parameters can not improve the performance. Our MEDOE architecture achieves the improvement by only increasing a few parameters. In general, the effectiveness of our method to be effective is dependent on the MEDOE strategy rather than the increase of parameters.

\begin{table*}[t]
\vspace{-0.5em}
    \centering
    \scalebox{0.9}{\begin{tabular}{l|c|c|c|c}
\hline \hline Methods & Backbone & mIoU~(\%) & mAcc~(\%) & Params~(GBs)  \\
\hline DeepLabv3+~\cite{r11-deeplab} & ResNet-50c & $80.37$ & $86.68$ & $24.16(1.0\times)$\\
 DeepLabv3+-MEDOE  & ResNet-50c &$80.62\color{green}({+0.25})$ & $90.04\color{green}({+3.36})$  &$34.31(1.4\times)$\\
 DeepLabv3+-MEDOE$^{\dagger}$  & ResNet-50c &$84.14\color{green}({+3.77})$ &$92.38\color{green}({+5.70})$  &$34.31(1.4\times)$\\
 PSPNet~\cite{r16-pspnet} & ResNet-50c &$78.32$ &$85.52$ &$18.33(1.0\times)$\\
 PSPNet-MEDOE & ResNet-50c &$78.82(\color{green}{+0.50})$ &$88.07(\color{green}{+2.55})$ &$20.37(1.1\times)$\\
 PSPNet-MEDOE$^{\dagger}$ & ResNet-50c &$84.32(\color{green}{+4.56})$ & $91.90(\color{green}{+5.34})$ &$20.37(1.1\times)$\\
 \hline
DeepLabv3+ & ResNet-101c & $80.67$ & $87.58$ &$25.02(1.0\times)$ \\
 DeepLabv3+-MEDOE & ResNet-101c & $81.21\color{green}({+0.54})$ & $91.29\color{green}({+3.69})$ &$35.03(1.4\times)$ \\
 DeepLabv3+-MEDOE$^{\dagger}$  & ResNet-101c &$84.51\color{green}({+3.84})$ &$92.63\color{green}({+5.05})$  &$35.03(1.4\times)$\\
 PSPNet & ResNet-101c &$79.76$ &$86.56$ &$20.11(1.0\times)$\\
 PSPNet-MEDOE & ResNet-101c &$79.79\color{green}({+0.03})$ &$90.88\color{green}({+4.32})$ &$23.81(1.1\times)$\\
 PSPNet-MEDOE$^{\dagger}$ & ResNet-101c &$84.32\color{green}({+4.56})$ & $91.90(\color{green}{+5.34})$ &$23.81(1.1\times)$\\
 \hline
OCRNet~\cite{r15-ocr} & HRNet-W48 & $80.70$ & $88.11$  &$23.44(1.0\times)$\\
 OCRNet-MEDOE & HRNet-W48 & $81.49\color{green}({+0.79})$ & $89.14\color{green}({+1.03})$ &$30.27(1.3\times)$\\
 OCRNet-MEDOE$^{\dagger}$ & HRNet-W48 & $85.29\color{green}({+4.59})$ & $93.38\color{green}({+5.27})$ &$30.27(1.3\times)$\\
 \hline
 SegFormer~\cite{xie2021segformer} &MIT-B3 & $81.94$ &$88.28$ &$30.02(1.0\times)$\\
 SegFormer-MEDOE &MIT-B3 &$82.37(\color{green}{+0.43})$ &$91.16(\color{green}{+2.88})$ &$38.65(1.3\times)$\\
  SegFormer-MEDOE$^{\dagger}$ &MIT-B3 &$85.67\color{green}({+3.73})$ &$93.49\color{green}({+5.21})$ &$38.65(1.3\times)$\\
\hline \hline
\end{tabular}
}
\caption{Comparison of performance and memory size~(Params) on the validation set of Cityscapes with current methods. $\dagger$: Oracle results for the ideal case.}
\vspace{-1em}
\label{paramt1}
\end{table*}

\begin{table}[t]
\setlength\tabcolsep{2.5pt}
    \scalebox{0.85}{\begin{tabular}{l|cc|cc|cc|cc}
        \hline \hline 
        \multirow{2}{*}{Method} & \multicolumn{2}{c|}{All} &
        \multicolumn{2}{c|}{Head} & \multicolumn{2}{c|}{Body} &
        \multicolumn{2}{c}{Few}\\
        & mIoU &mAcc & mIoU &mAcc & mIoU &mAcc & mIoU &mAcc\\
        \hline Baseline & $42.11$ & $54.13$ & $65.50$ & $\textbf{78.71}$ & $44.75$ &$59.75$ & $33.64$ &$42.97$ \\ 
        w/ Overlap &$\textbf{43.82}$ & $\textbf{60.02}$ & $\textbf{66.99}$ &$77.90$ &$\textbf{46.32}$ &$\textbf{63.52}$ &$\textbf{35.38}$ &$\textbf{52.33}$ \\
        w/o Overlap & $43.18$ & $58.91$ & $66.43$ & $77.64$ & $45.49$ &$63.44$ & $34.52$ &$48.95$ \\
        \hline \hline
    \end{tabular}}
    \vspace{-0.3em}
    \captionof{table}{Ablation of overlapping strategy with DeepLabv3+ \& ResNet-50c on ADE20K.}
    \label{overlap}
\vspace{-1.5em}
\end{table}
\vspace{-0.5em}

\begin{table*}[t]
    \centering
    \scalebox{0.9}{\begin{tabular}{l|c|c|c|c}
\hline \hline Methods & Backbone & mIoU~(\%) & mAcc~(\%) & Params~(GBs)  \\
\hline DeepLabv3+~\cite{r11-deeplab} & ResNet-50c & $42.11$ & $54.13$ & $25.82(1.0\times)$\\
 DeepLabv3+-MEDOE  & ResNet-50c &$43.82(\color{green}{+1.71})$ & $60.02(\color{green}{+5.89})$  &$35.49(1.4\times)$\\
 DeepLabv3+-MEDOE$^{\dagger}$ & ResNet-50c & $53.34 (\color{green}{+11.23}) $
 & $73.97 (\color{green}{+19.84}) $  &$35.49(1.4\times)$\\
 PSPNet~\cite{r16-pspnet} & ResNet-50c &$40.46$ &$51.42$ &$19.81(1.0\times)$\\
 PSPNet-MEDOE & ResNet-50c &$41.76(\color{green}{+1.30})$ &$54.27(\color{green}{+2.85})$ &$23.49(1.1\times)$\\
 PSPNet-MEDOE$^{\dagger}$ & ResNet-50c &$52.00(\color{green}{+11.54})$ & $71.61(\color{green}{+20.19})$ &$23.49(1.1\times)$\\
 \hline
DeepLabv3+ & ResNet-101c & $44.60$ & $56.28$ &$26.31(1.0\times)$ \\
 DeepLabv3+-MEDOE & ResNet-101c & $46.13(\color{green}{+1.42})$ & $61.12(\color{green}{+4.84})$  &$36.27(1.4\times)$\\
 DeepLabv3+-MEDOE$^{\dagger}$  & ResNet-101c &$55.18(\color{green}{+10.58})$ &$76.82(\color{green}{+20.54})$  &$36.27(1.4\times)$\\
 PSPNet & ResNet-101c &$43.33$ &$54.51$ &$22.94(1.0\times)$\\
 PSPNet-MEDOE & ResNet-101c &$44.31(\color{green}{+0.98})$ &$59.85(\color{green}{+5.19})$ &$25.69(1.1\times)$\\
 PSPNet-MEDOE$^{\dagger}$ & ResNet-101c &$53.47(\color{green}{+10.14})$ & $72.86(\color{green}{+18.45})$ &$25.69(1.1\times)$\\
 \hline
OCRNet~\cite{r15-ocr} & HRNet-W48 & $42.53$ & $54.91$  &$33.03(1.0\times)$\\
 OCRNet-MEDOE & HRNet-W48 & $43.31(\color{green}{+0.78})$ & $58.96(\color{green}{+4.05})$ &$37.67(1.1\times)$\\
 OCRNet-MEDOE$^{\dagger}$ & HRNet-W48 & $51.99(\color{green}{+9.46})$ & $74.68(\color{green}{+19.77})$ &$37.67(1.1\times)$\\
 \hline
 SegFormer~\cite{xie2021segformer} &MIT-B3 & $47.13$ &$60.84$ &$30.02(1.0\times)$\\
 SegFormer-MEDOE &MIT-B3 &$48.22(\color{green}{+1.09})$ &$64.06(\color{green}{+3.22})$ &$33.96(1.3\times)$\\
  SegFormer-MEDOE$^{\dagger}$ &MIT-B3 &$56.49(\color{green}{+9.36})$ &$79.41(\color{green}{+18.57})$ &$33.96(1.3\times)$\\
 
\hline \hline
\end{tabular}
}
\caption{Comparison of performance and memory size~(Params) on the validation set of ADE20K with with current methods. $\dagger$: Oracle results for the ideal case.}
\label{paramt2}
\vspace{-1em}
\end{table*}


\begin{table}[t]
    \centering
\scalebox{1}{\begin{tabular}{c|c}
        \hline \hline Dataset &$\rho_{X,Y}(\%)$  \\
        \hline 
        CIFAR-100 & $75.9$ \\
        ADE20K-pixel & $36.8$ \\
        Cityscapes-pixel & $58.9$  \\
        \hline \hline
    \end{tabular}
    }
    \caption{ Pearson coefficients between categories frequency and
accuracy, the lower value means a much weaker correlation. }
\label{t7}
\vspace{-1em}
\end{table}
\subsection{Proof of the Re-balancing Methods' Weakness}

In this section, to verify the internal reason of the gap between long-tailed semantic segmentation and long-tailed recognition of re-balancing method. We calculate the Pearson correlation coefficient to show the correlation between category accuracy and category frequency (image category frequency on CIFAR-100) in Table~\ref{t7}. The weak correlation between category accuracy and pixel level frequency on Cityscapes and ADE20K cause challenges to re-weighting in semantic segmentation. which can be demonstrated in Table~\ref{t8}. In general, the current Pixel level re-balancing approaches can not work well with the long tail distribution on semantic segmentation.

\subsection{Analysis of the performance gap between Cityscapes and ADE20K} \label{A.5}

According to the experiments' results of Sec~\ref{dataset}, our method has achieved impressive performance in both mIoU and mAcc on Cityscapes and ADE20K dataset. However, there seems to be a gap between the performance on Cityscapes and ADE20K. We analyze the causes: 1) Compared to ADE20K, there are fewer categories in Cityscapes and a more pronounced long-tailed distribution (higher proportion of head category instances), so it is easier to fall into local optimum when training middle and tail experts, resulting in the  overfitting of these categories. Finally, it caused that the increase of \(\textit{FN}_{ht}\) and \(\textit{FP}_{ht}\) on the overall datasets. \(\textit{FN}_{ij}\) and \(\textit{FP}_{ij}\) denote to the pixel instance in which the ground truth belongs to category \(\textit{i}\) but the prediction is \(\textit{j}\). 2) The performance of baseline methods on Cityscapes is at a high level, espically, for the head categories. According to the above two reasons, our method improves the overall \(\textit{TP}\) on Cityscapes, but it will increase \(\textit{FN}_{ht}\) of the head categories and \(\textit{FP}_{ht}\) of the tail categories, respectively, resulting in \(\textit{mIoU}\) being at a relatively stable value. 

As shown in Figure~\ref{fig.vis_all}, our method misclassified limited pixels of head categories, which surrounds the pixels of tail categories, as tail categories. Given the experimental results in Table~\ref{t_orcal}, such a compromise that can increase the segmentation performance for tail categories is supposed to be reasonable and acceptable in real-world applications.

\subsection{Additional Explanations}

\paragraph{Contextual Module} as an important module in semantic segmentation, refers to the extraction and aggregation of contextual information for pixels through a series of operations (\ie, feature pyramids, atrous convolution, large-scale convolutional, attention mechanisms, or global pooling). Existing contextual modules include: ASPP~\cite{r11-deeplab}, PSP~\cite{r16-pspnet}, or non-local~\cite{DBLP:journals/corr/abs-1711-07971}.

\paragraph{Contextual Information} means the relationship between this pixel and the surrounding pixels and global information is regarded as contextual information. \textbf{The reasons why segmentation needs contextual module.} When solving semantic segmentation tasks if each pixel considers only its deep features, such as texture and color, it will be difficult to classify into the correct category (\ie the deep features of leaves and green grass will be very similar). At the same time, as stated in the work, semantic segmentation believes that each pixel is not I.I.D., but is related to the surrounding pixels. The correlation information can better help the segmentation task (\ie the leaves are surrounded by branches, and the grass is likely to be surrounded by roads). 

\subsection{Ablation of Different Experts}

To measure the goodness and distinctiveness, We adopt the mAcc and per-category bias in Table \ref{t9} as the metrics to describe each expert. It seems to demonstrate that the \textit{K} experts learned at Stage 1 are good and distinctive from each other, rather than simply boosting confidence through multi-expert training.

Furthermore, according to the well-known bias-variance decomposition~\cite{r24-ride}, per-category bias denotes: 
\begin{equation}
\begin{aligned}
& \operatorname{Error}(x ; h) = E\left[(h(x ; D)-Y)^{2}\right] \\
& = \operatorname{Bias}(x ; h)+\operatorname{Variance}(x ; h)+\text { irreducible error }(x),
\label{11}
\end{aligned}
\end{equation}

\begin{table}[ht]
    \centering
    \scalebox{1}{\begin{tabular}{l|cc|cc|cc}
        \hline \hline 
        \multicolumn{1}{l|}{} & \multicolumn{2}{c|}{Head} & \multicolumn{2}{c|}{Body} &
        \multicolumn{2}{c}{Few}\\
        & mAcc &bias &mAcc &bias &mAcc &bias \\
        \hline Overall & $0.96$ & $0.10$ & $0.88$ &$0.22$ & $0.81$ &$0.30$\\
        \hline Expert 1 & $0.97$ & $0.09$ & $0.86$ &$0.24$ &$0.75$ &$0.36$\\
        Expert 2 &- &- &$0.95$ & $0.09$ & $0.86$ & $0.19$\\
        Expert 3 &-&-&-&- &$0.91$ & $0.13$  \\
        \hline \hline
    \end{tabular}
    }
    \caption{Ablation of each expert with the mAcc and per-category bias on Cityscapes with MEDOE (DeepLabv3+ \& ResNet-50c). The higher mAcc and lower bias are better.}
    \vspace{-1em}
    \label{t9}
\end{table}

\subsection{Theoretical Analysis on Metrics \label{theoretical}}
Given the experimental results on Cityscapes, it is surprising to find that the increase in mAcc is much higher than that in the mIoU. This misalignment inspires us to explore the relationship between mIoU and mAcc in long-tailed semantic segmentation, for we employed mIoU as the primary metric for evaluation, whereas studies on long-tailed recognition preferred mAcc. Suppose true positives, false negatives, and false positives are denoted by $TP$, $FN$, and $FP$, respectively, pixel classification accuracy (Acc) and Intersection-over-Union (IoU) for category \textit{i} are formulated as follows,
\begin{equation}
\begin{aligned}
A c c_{i}&=\frac{T P_{i}}{({T P_{i}+F N_{i}})}, \\
I o U_{i}&=\frac{T P_{i}}{({T P_{i}+F N_{i}+F P_{i}})}.
\label{eq:8}
\end{aligned}
\end{equation}
As $\sum^{c}_{i=1}{FP}_{i} = \sum^{c}_{i=1}{FN}_{i}$ and $\sum^{c}_{i=1}{FN}_{i}$ is more affected by the incorrect instances for head categories than for tail categories (See Remark1 for details), the $FN$-related $FP$ can be suppressed by head categories while stabilizing IoU in the evaluation in IoU for tail categories. 

We regard mIoU as a metric focusing on the overall performance for segmenting an image, on which head categories largely impact, whereas mAcc is more category-sensitive and should be of higher importance in our experiments to evaluate the segmentation performance for body and tail categories. In addition, we observe that pixels nearby the instances of a tail category $i$ were likely to be classified as $i$, which resulted in a substantial increase in $FP_{i}$ and jeopardized the segmentation performance. Therefore, the relationship between mIoU and mAcc can be summarized as that our observed jeopardy is less likely to occur upon an increase in mAcc unless mIoU significantly drops (See Remark2 for details).

\subsection{Additional Proof of Theoretical Analysis between MIoU and MAcc}

Before we proceed with the additional proof, we introduce some formulas for mAcc and mIoU.
\begin{equation}
\begin{gathered}
{m A c c}= \frac{1}{C}\sum^{C}_{i=1}{Acc}_{i} =\frac{1}{C}\sum^{C}_{i=1}\frac{T P_{i}}{T P_{i}+F N_{i}}, \\
{m I o U}=\frac{1}{C}\sum^{C}_{i=1}{IoU}_{i} =\frac{1}{C}\sum^{C}_{i=1}\frac{T P_{i}}{T P_{i}+F N_{i}+F P_{i}}.
\end{gathered}
\label{eq:miou}
\end{equation}

\begin{equation}
    \sum^{C}_{i=1}{FP}_{i}=\sum^{C}_{i=1}{FN}_{i} \& {n u m_{i}} = {T P}_{i} + {F P}_{i}
    \label{eq.fnfp}
\end{equation}


\paragraph{Remark1:}According to the prior information from experimental statistics, the ratios of pixel instances of the head, body, and tail categories in the semantic segmentation are 80\%, 15\%, and 5\%.
We analyzed the items in the mIoU and mAcc formulas and found that the difference between them was the $\mathit{FP}_{i}$. According to Eq.~\eqref{eq.fnfp} We derive Eq.~\eqref{eq:FN}. Where, \textit{h}, \textit{b} and \textit{t} refer to head, body, tail categoris and $\mathit{NUM}$ refers to the pixel instance number of overall datasets.

From Eq.~\eqref{eq:FN}, we obtain the conclusion that due to the large instance base of the head categories, \(\textit{FN}_{i}\) and \textit{FN}-related \(\textit{FP}_{i}\) items are dominated by head categories. Therefore, the \(\textit{IoU}_{i}\) of each category \textit{i} and \(\textit{mIoU}\) will be dominated by head categories because of the item \(\textit{FP}_{i}\). 

Instead of \(\textit{mIoU}\), for each category \textit{i} of \(\textit{mAcc}\), the items of \(\textit{Acc}_{i}\) are only related to its own category \textit{i}, and \(\textit{mAcc}\) will not be dominated by the head categories. Thus \(\textit{mAcc}\) is a fair and tail-sensitive metric.


\paragraph{Remark2:}To better understand the correlation between mean IoU and mean Acc in long-tailed semantic segmentation. According to the equation Eq.~\eqref{eq:8} and the results from the experiments. The category Acc and IoU in our method become:
\begin{equation}
\begin{gathered}
\hat{I o U_{i}}  \approx {I o U_{i}}, \\
\hat{A c c_{i}} = ( 1 + p) {A C C}_{i}.
\end{gathered}
\label{eq:12}
\end{equation}
And it is clear from Eq.~\eqref{eq.fnfp}, We obtain the result in Eq.~\eqref{eq.15}:

\begin{strip}
\begin{equation}
\begin{gathered}
    \textstyle{\sum}^{C}_{i=1}{FP}_{i} = \textstyle{\sum}^{C}_{i=1}{FN}_{i} = \textstyle{\sum}^{{C}_{h}}{FN}_{i} + \textstyle{\sum}^{{C}_{b}}{FN}_{i} + \textstyle{\sum}^{{C}_{t}}{FN}_{i} \\
    = \textstyle{\sum}^{{C}_{h}}(1-{Acc}_{i})\times {num}_{i} + \textstyle{\sum}^{{C}_{b}}(1-{Acc}_{i})\times {num}_{i} + \textstyle{\sum}^{{C}_{t}}(1-{Acc}_{i})\times {num}_{i} \\
    \approx{} (1-{mAcc}_{h}) \textstyle{\sum}^{{C}_{h}} {num}_{i} + (1-{mAcc}_{b}) \textstyle{\sum}^{{C}_{b}}{num}_{i} + (1-{mAcc}_{t}) \textstyle{\sum}^{{C}_{t}} {num}_{i} \\
    = (1-{mAcc}_{h}) \times 0.8{NUM} + (1-{mAcc}_{b}) \times 0.15{NUM} + (1-{mAcc}_{t}) \times 0.05{NUM} 
    \end{gathered}
\label{eq:FN}
\end{equation}
\begin{equation}
    \begin{gathered}
    \hat{A c c_{i}} = ( 1 + p) {A C C}_{i}\\
    \Rightarrow{}
    \hat{A c c_{i}} = \frac{\hat{T P_{i}}}{\hat{T P_{i}}+\hat{F N_{i}}} = \frac{\hat{T P_{i}}}{T P_{i}+F N_{i}} = (1+p) \frac{T P_{i}}{T P_{i}+F N_{i}} = ( 1 + p) {A C C}_{i} \\
    \Rightarrow{}
    \hat{T P_{i}} = ( 1 + p) {T P_{i}}
    \end{gathered}
\end{equation}

\begin{equation}
    \begin{gathered}
    \hat{I o U_{i}}  \approx {I o U_{i}}
    \Rightarrow{}
    \hat{I o U_{i}} =  \frac{\hat{T P_{i}}}{\hat{T P_{i}}+\hat{F N_{i}} + \hat{{F P}_{i}}} = \frac{\hat{T P_{i}}}{T P_{i}+F N_{i} + \hat{{F P}_{i}}} \approx{} \frac{T P_{i}}{T P_{i}+F N_{i}+F P_{i}} = I o U_{i} \\
    \Rightarrow{}
    (1+p){T P}_{i} \times ({T P_{i}+F N_{i}+F P_{i}}) = {T P}_{i} \times ({T P_{i}+F N_{i} + \hat{{F P}_{i}}}) \\
    \Rightarrow{}
    (1+p){T P}_{i} \times ({{n u m}_{i}+F P_{i}}) = {T P}_{i} \times ({{n u m}_{i} + \hat{{F P}_{i}}}) \\
    \Rightarrow{}
    \Delta{F P_{i}} = p \times ({n u m}_{i} + {F P}_{i})
    \end{gathered}
    \label{eq.15}
\end{equation}
\begin{equation}
    \begin{gathered}
    A c c_{i}=\frac{T P_{i}}{T P_{i}+F N_{i}} \ \& \ 
I o U_{i}=\frac{T P_{i}}{T P_{i}+F N_{i}+F P_{i}}
    \Rightarrow{}
    {F P }_{i} = \frac{Acc_{i}}{IoU_{i}} \times {num}_{i} - {num}_{i}
    \end{gathered}
    \label{eq.16}
\end{equation}
\end{strip}

where \(\Delta{F P_{i}}\) is the increased false positive from baseline to our method and $p$ is the improvement percent in $Acc_i$. 

To guarantee the classifier is effective and segment more tail categories, it should satisfy:
\begin{equation}
    \begin{gathered}
    \Delta{F P_{i}} = p \times ({n u m}_{i} + {F P}_{i}) \ll num_{i}
    \end{gathered}
    \label{17}
\end{equation}

We observe that the results in the Cityscapes benchmark have achieved a high value, \eg  ${IoU}_{i}$ = 0.8 and ${Acc}_{i}$ = 0.85. According to this precondition and Eq.~\eqref{eq.16}, ${F P}_{i}$ = 0.0625 $\times {n u m}_{i}$ and $\mathit{p}$ are both small value, which means $\Delta{F P_{i}}$ is a minimum value and satisfied Eq.~\eqref{17}. This is just the reason that Acc improved while IoU does not decrease significantly for tail categories segmentation.

\subsection{Contemporaneous Long-tailed semantic segmentation}
    To the best of our knowledge, there is a contemporaneous work, Region rebalance~\cite{r32-RR}, with our paper. Although, both of us almost simultaneously focus on the long-tailed distribution as an important reason for constraining semantic segmentation performance, there exist differences between our work and Region Rebalance~\cite{r32-RR}: 
    
    \paragraph{About the Perspective:} Region Rebalance was concerned about solving the problem of category rebalance, while our work is more focused on improving the recognition of tail and body categories and placed special emphasis on the significance of segmenting the body, and tail categories.
    
    \paragraph{About the Method:} Region Rebalance relieves the categories imbalance with an auxiliary region classification branch by adjusting segmentation boundaries. However motivated by the ensemble-and-grouping methods, we proposed MEDOE framework to encourage different experts to learn more balanced distribution in the feature space, and finally adjust classification boundaries for each experts. Compared to Region Rebalance, the motivation of our work was completely different and provided different research directions.
    
    \paragraph{About the discussion on mIoU and mAcc:} Region Rebalance only explains the cause of mIoU with previous long-tailed methods. Different from Region rebalance, we take a further step and explore the significance of mAcc in segmenting the body and tail categories. 
    
    \paragraph{About the dataset:}Region Rebalance uses COCO164 (164 categories) and ADE20K (150 categories) datasets, which both comprise massive categories. In addition to ADE20K, we use Cityscapes (19 categories) to further evaluate the case with less categories. The performance gap between Cityscapes and COCO164 may be partly due to the impact of the dataset, we also demonstrated the gap between Cityscapes and ADE20K in Appendix~\ref{A.5}. Our method demonstrate excellent results on a large number of categories of datasets such as ADE20K. 

    \paragraph{About the setting:}Region rebalance is only experimented with the current semantic segmentation dataset setting.  However, referring to the setting of long-tail recognition tasks, we conduct experiments under the setting of uniform distribution semantic segmentation.  As shown in Appendix~\ref{uniform_set}, our method is fairer for the body and tail categories.

\subsection{Significance of Focusing on Long-tailed Semantic Segmentation in Real-World Applications}
In this section, we will illustrate the significance of focusing on long-tailed methods in real-world semantic segmentation task.
\paragraph{Significance in more data distributions}As we mentioned in the introduction, taking into account the variation of real-world data distribution, (\ie, more person during rush hours than normal times). The long-tailed semantic segmentation methods are fairer to the tail and body categories and are robust to the change in data distribution by balancing the contribution for each category.  

\paragraph{Significance of segmenting body and tail categories in real-world}We believe in a large number of real-world applications, it is more important to be able to identify body and tail categories than accurately segment the edge pixels of head categories. (\ie for autonomous vehicles, it is necessary to segment some tail categories objects that appear on the driving path, such as ``poles" or ``fire hydrants", to avoid traffic accidents. Segmenting small lesions on medical images can help doctors detect underlying diseases). Generally, the benefits of this segmentation are far greater than the decrease of certain head categories edges. 

\subsection{Comparison with Traditional Multi-Experts Methods}
We compared the differences from existing multi-expert methods, there is three main difference: 

\paragraph{Label Masking Strategy} Previous ensemble-and-grouping methods mostly adopt some re-weighting and re-sampling strategies to help the body- and tail-expert learn the dominant categories. However, as mentioned in \ref{intro}, the re-weighting strategies are not applicable because the tight correlation between image pixels and the long-tailed degree of semantic segmentation at the pixel level is much higher than classification tasks (the ratio of head- to tail-categories). And simply re-sampling strategies will undermine the contextual information of an image. Therefore, we propose a masking strategy for training labels, \emph{expert-specific pixel-masking strategy}. A re-sampling-like strategy, which can preserve the context information by masking the head-categories labels, head- and body-categories labels to assist the training of body- and tail-experts, respectively.

\paragraph{Model Architecture} The pipeline of traditional multi-expert methods contains a backbone and multi-classifiers to adjust classifier boundaries and finally ensemble the outputs. 1)We pioneered the \textbf{combination of contextual modules and classifiers} to become experts and learn more balanced distribution in the feature space, and finally adjust classification boundaries. 2)Then we provided each expert with soft weight based on the final contextual information and classification results through a learning mechanism to ensemble the outputs.

\paragraph{Training Strategies}Traditional methods often focus on constraining dominant categories and ignoring the confusing categories. Differing from them, we proposed the expert-specific pixel-masking strategy and diverse distribution-aware loss function to ensure our model architecture focuses on the confusing categories and has better performance.
\paragraph{Training Step}Advanced multi-expert methods in long-tailed classification, such as BBN~\cite{r37-bbn}, RIDE~\cite{r24-ride}, LFME~\cite{r35-lfme}. They all take multiple steps to train, mainly including 1. training backbone, 2. training classifiers, 3. Distillation learning (optional). However, our method can update the backbone parameters while training the head expert.

\subsection{Quantitative Visualization Comparisons on Cityscapes and ADE20K }

\begin{figure*}[!t]
\label{viso_city}
    \centering
    \subfigure[Image]{
    \begin{minipage}[t]{0.22\linewidth}
    \centering
    \includegraphics[width = 1.25in]{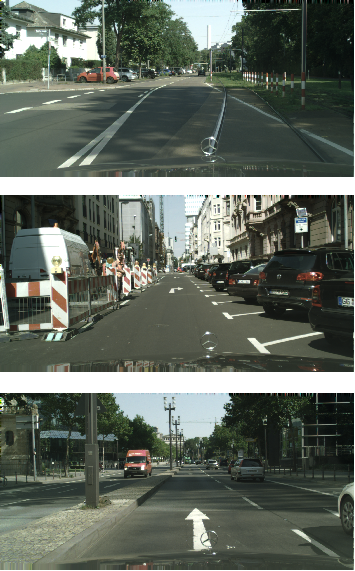}
    \end{minipage}}
    \subfigure[Ground Truth]{
    \begin{minipage}[t]{0.22\linewidth}
    \centering
    \includegraphics[width = 1.25in]{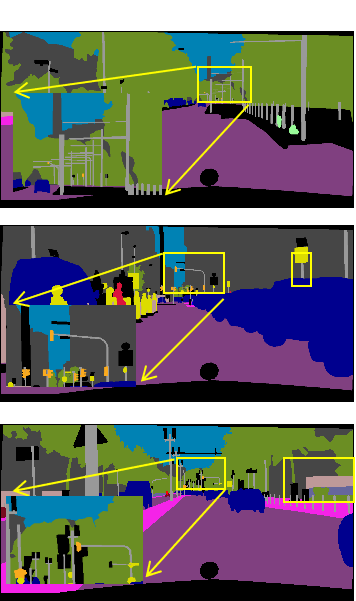}
    \end{minipage}}
    \subfigure[DeepLabv3+]{
    \begin{minipage}[t]{0.22\linewidth}
    \centering
    \includegraphics[width = 1.25in]{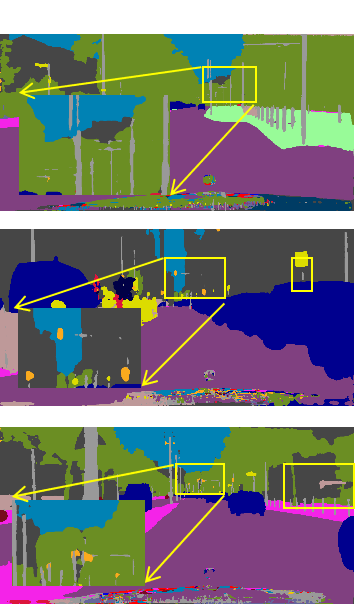}
    \end{minipage}}
    \subfigure[MEDOE~(Ours)]{
    \begin{minipage}[t]{0.22\linewidth}
    \centering
    \includegraphics[width = 1.25in]{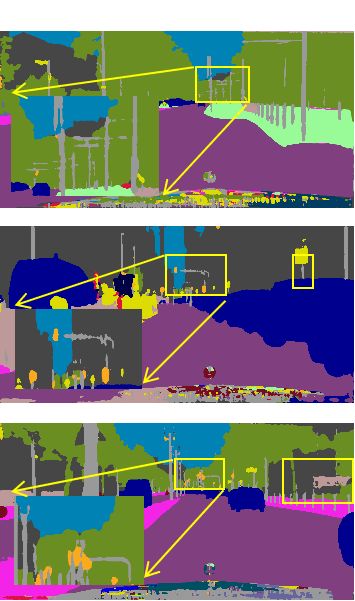}
    \end{minipage}}
    \caption{Qualitative Visualization results on the validation set of Cityscapes with ResNet-50c as the backbone. All the models here are trained under the same setting. }
    \label{fig.vis_all}
\end{figure*}

\begin{figure*}[!t]
\label{viso_ade2}
    \centering
    \subfigure[Image]{
    \begin{minipage}[t]{0.22\linewidth}
    \centering
    \includegraphics[width = 1.21in]{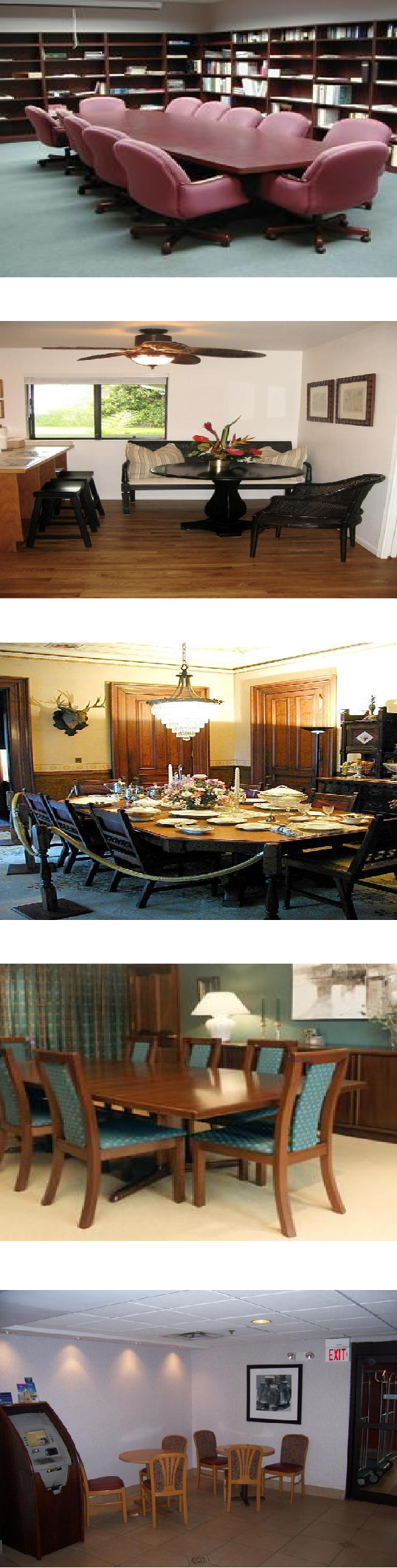}
    \end{minipage}}
    \subfigure[Ground Truth]{
    \begin{minipage}[t]{0.22\linewidth}
    \centering
    \includegraphics[width = 1.21in]{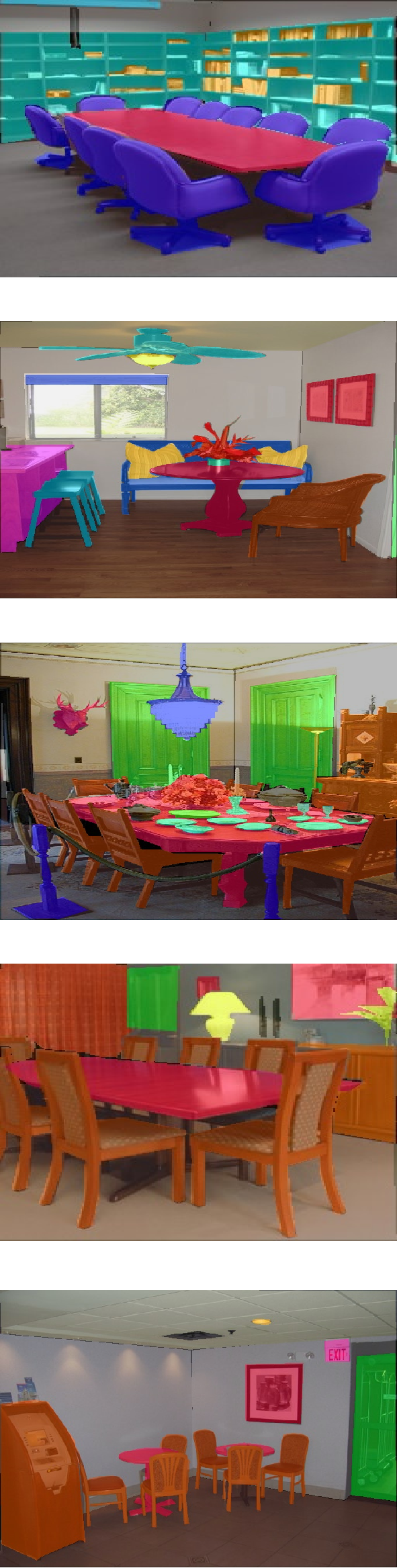}
    \end{minipage}}
    \subfigure[DeepLabv3+]{
    \begin{minipage}[t]{0.22\linewidth}
    \centering
    \includegraphics[width = 1.21in]{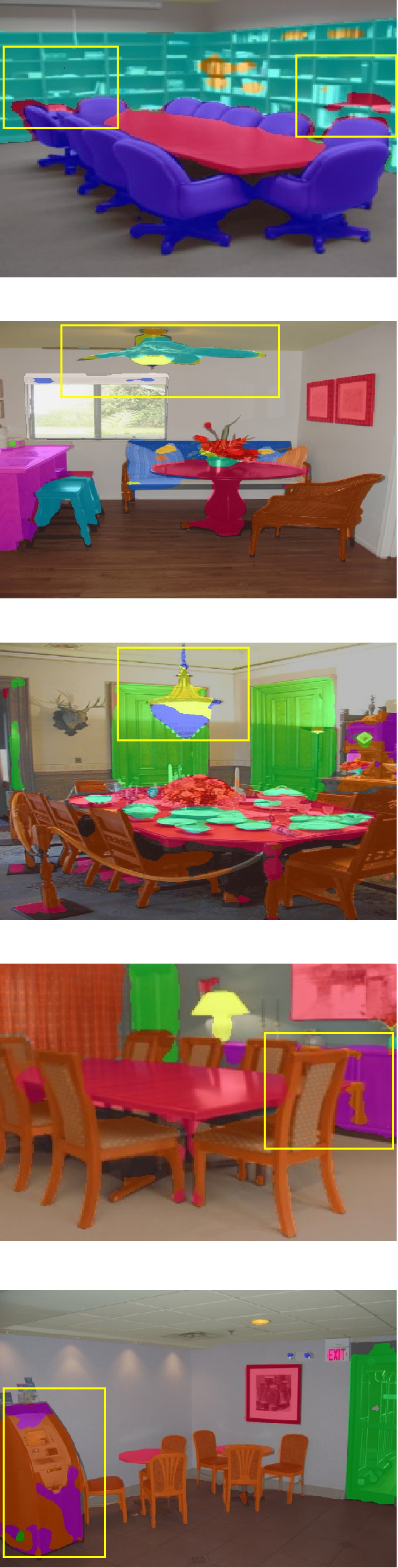}
    \end{minipage}}
    \subfigure[MEDOE~(Ours)]{
    \begin{minipage}[t]{0.22\linewidth}
    \centering
    \includegraphics[width = 1.21in]{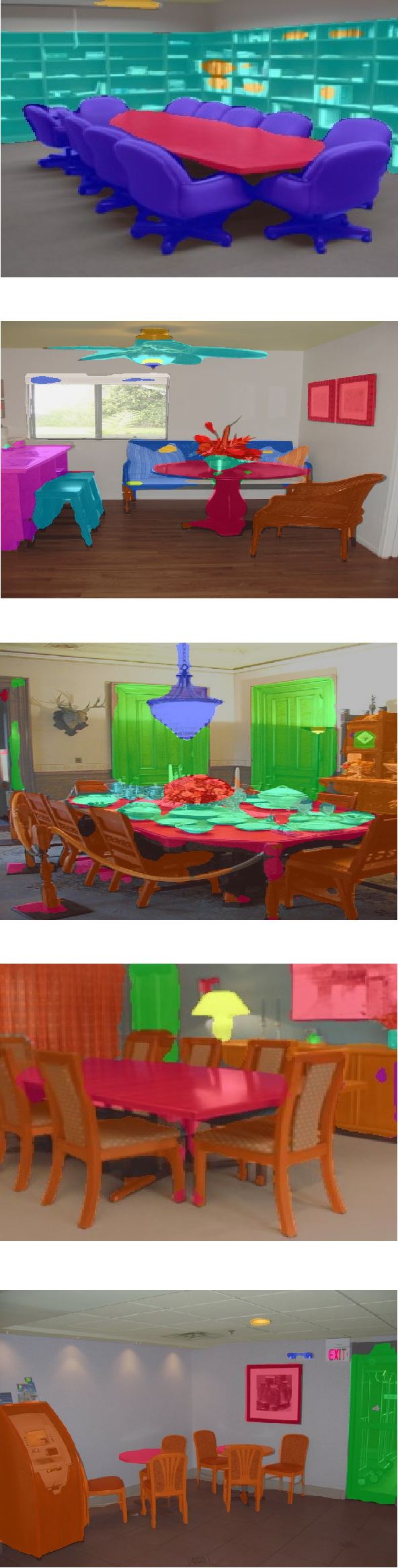}
    \end{minipage}}
    \caption{Qualitative Visualization on the validation set of ADE20K with ResNet-50c as the backbone. All the models here are trained under the same setting. }
    \label{fig.ade_vis2}
\end{figure*}

In this section, we demonstrate the better performance of the MEDOE framework with quantitative visualization on Cityscapes and ADE20K shown in Figure~\ref{fig.vis_all} and~\ref{fig.ade_vis2}. We adopt ResNet-50c as the backbone and all models are trained under the same setting. In extensive semantic segmentation tasks, our MEDOE method can achieve better performance in segmenting tail categories.

\end{document}